\documentclass[runningheads]{llncs}

\usepackage{eccv}

\usepackage{eccvabbrv}

\usepackage{graphicx}
\usepackage{booktabs}
\usepackage{array} 
\usepackage{adjustbox} 

\usepackage[accsupp]{axessibility}  

\usepackage{hyperref}

\usepackage{orcidlink}

\begin{document}

\title{Fast Context-Based Low-Light Image Enhancement via Neural Implicit Representations} 

\titlerunning{Fast Context-Based Low-Light Image Enhancement via NIRs}

\author{Tomáš Chobola\inst{1,2}\orcidlink{0009-0000-3272-9996} \and
Yu Liu\inst{1}\orcidlink{0000-0003-2281-6791} \and
Hanyi Zhang\inst{1,2}\orcidlink{0009-0001-6314-9158} \and
Julia A. Schnabel\inst{1,2,3}\orcidlink{0000-0001-6107-3009} \and
Tingying Peng\inst{1,2}\orcidlink{0000-0002-7881-1749}
}

\authorrunning{T. Chobola et al.}

\institute{School of Computation, Information and Technology, Technical University of Munich, Munich, Germany \and Helmholtz AI, Helmholtz Munich - German Research Center for Environmental Health, Neuherberg, Germany \and School of Biomedical Engineering and Imaging Sciences, King’s College London, London, UK}

\maketitle

\begin{abstract}
  Current deep learning-based low-light image enhancement methods often struggle with high-resolution images, and fail to meet the practical demands of visual perception across diverse and unseen scenarios. In this paper, we introduce a novel approach termed \textit{CoLIE}, which redefines the enhancement process through mapping the 2D coordinates of an underexposed image to its illumination component, conditioned on local context. We propose a reconstruction of enhanced-light images within the HSV space utilizing an implicit neural function combined with an embedded guided filter, thereby significantly reducing computational overhead. Moreover, we introduce a single image-based training loss function to enhance the model's adaptability to various scenes, further enhancing its practical applicability. Through rigorous evaluations, we analyze the properties of our proposed framework, demonstrating its superiority in both image quality and scene adaptability. Furthermore, our evaluation extends to applications in downstream tasks within low-light scenarios, underscoring the practical utility of \textit{CoLIE}. The source code is available at \href{https://github.com/ctom2/colie}{https://github.com/ctom2/colie}.

\end{abstract}

\section{Introduction}
\label{sec:introduction}

\begin{figure}[t]
    \begin{subfigure}[b]{0.5\textwidth}
        \centering
        \includegraphics[width=.7\linewidth]{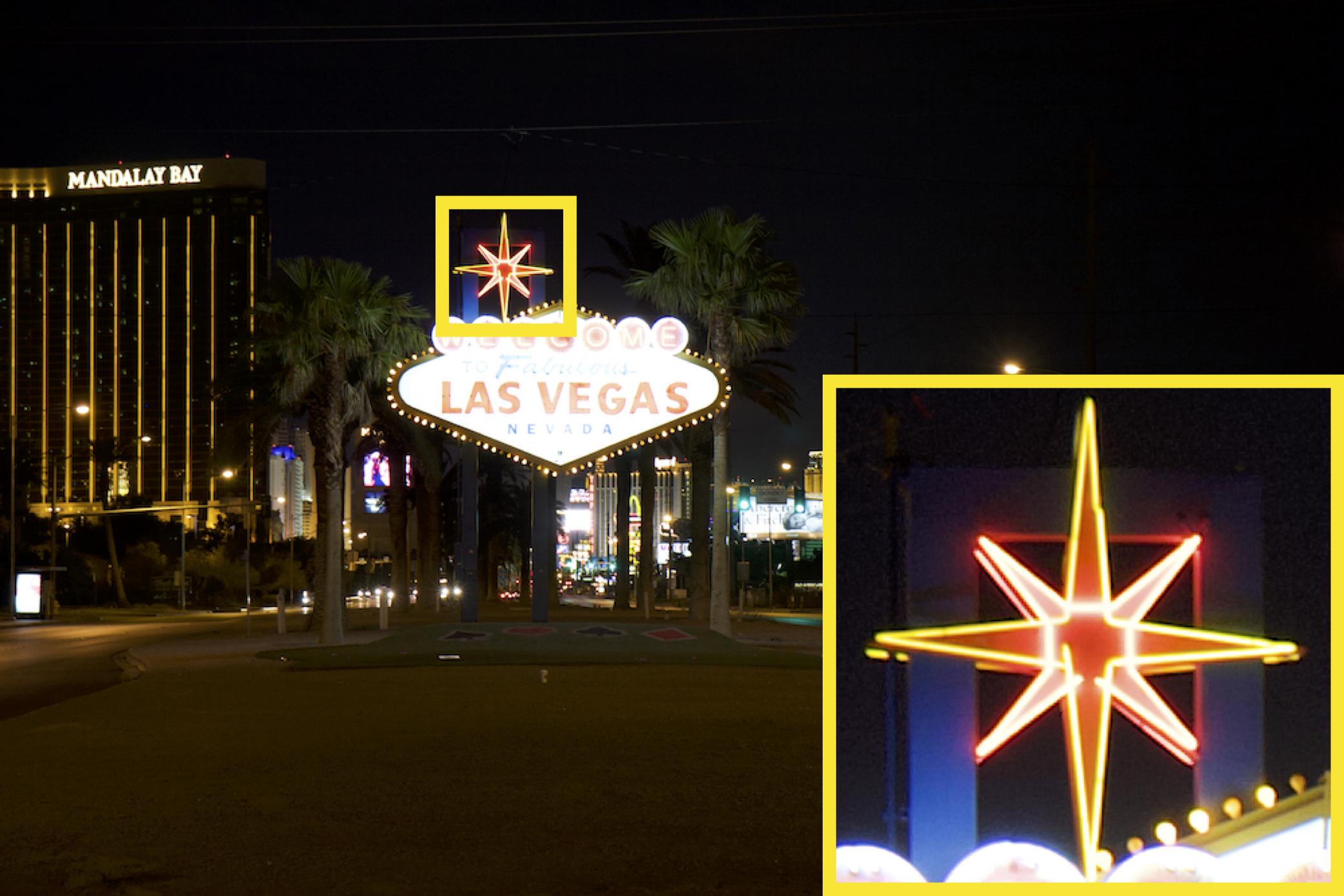}
        \caption{Input}
    \end{subfigure}%
    \begin{subfigure}[b]{0.5\textwidth}
        \centering
        \includegraphics[width=.7\linewidth]{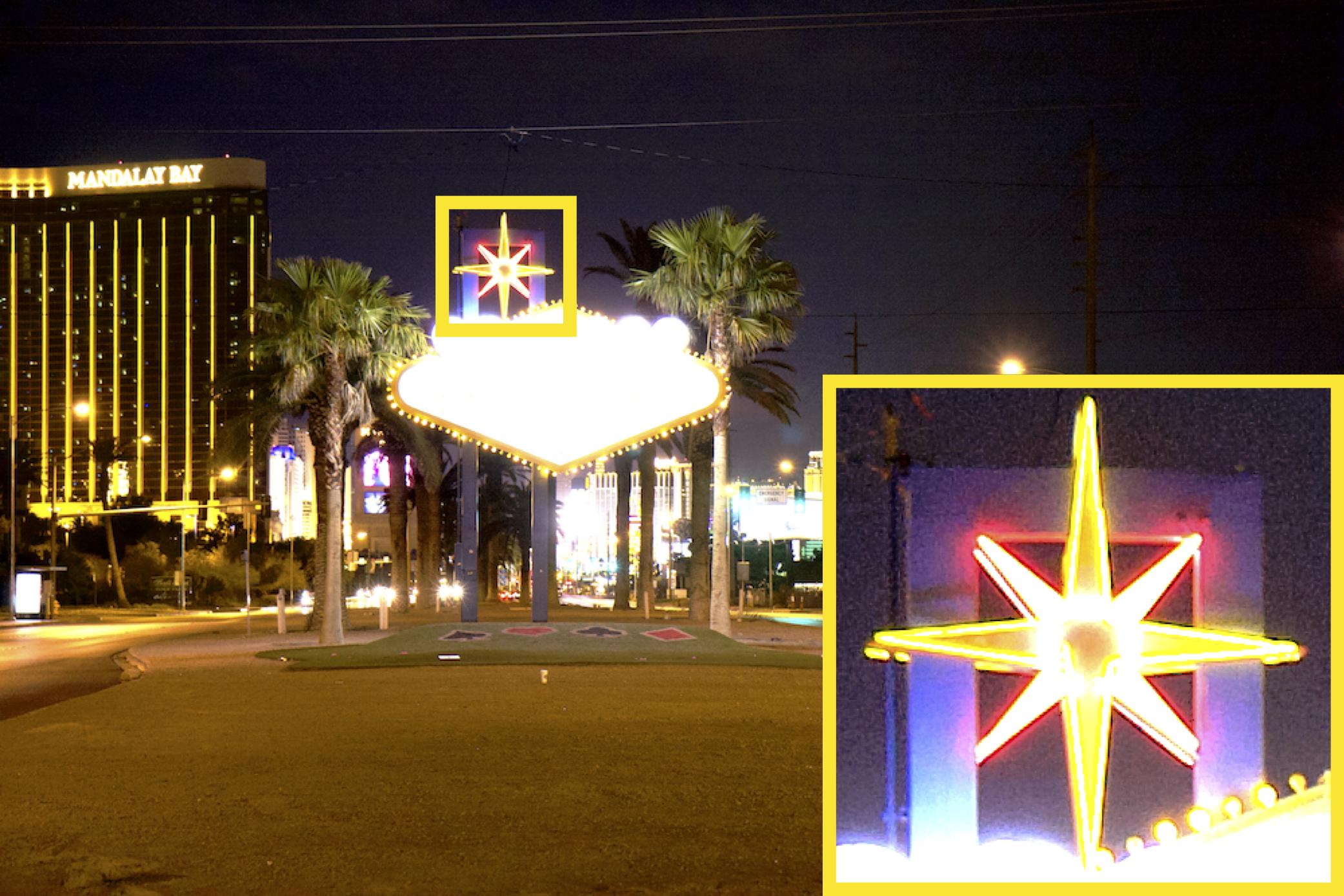}
        \caption{RUAS \cite{liu2021retinex}}
    \end{subfigure}%
    \newline
    \begin{subfigure}[b]{0.5\textwidth}
        \centering
        \includegraphics[width=.7\linewidth]{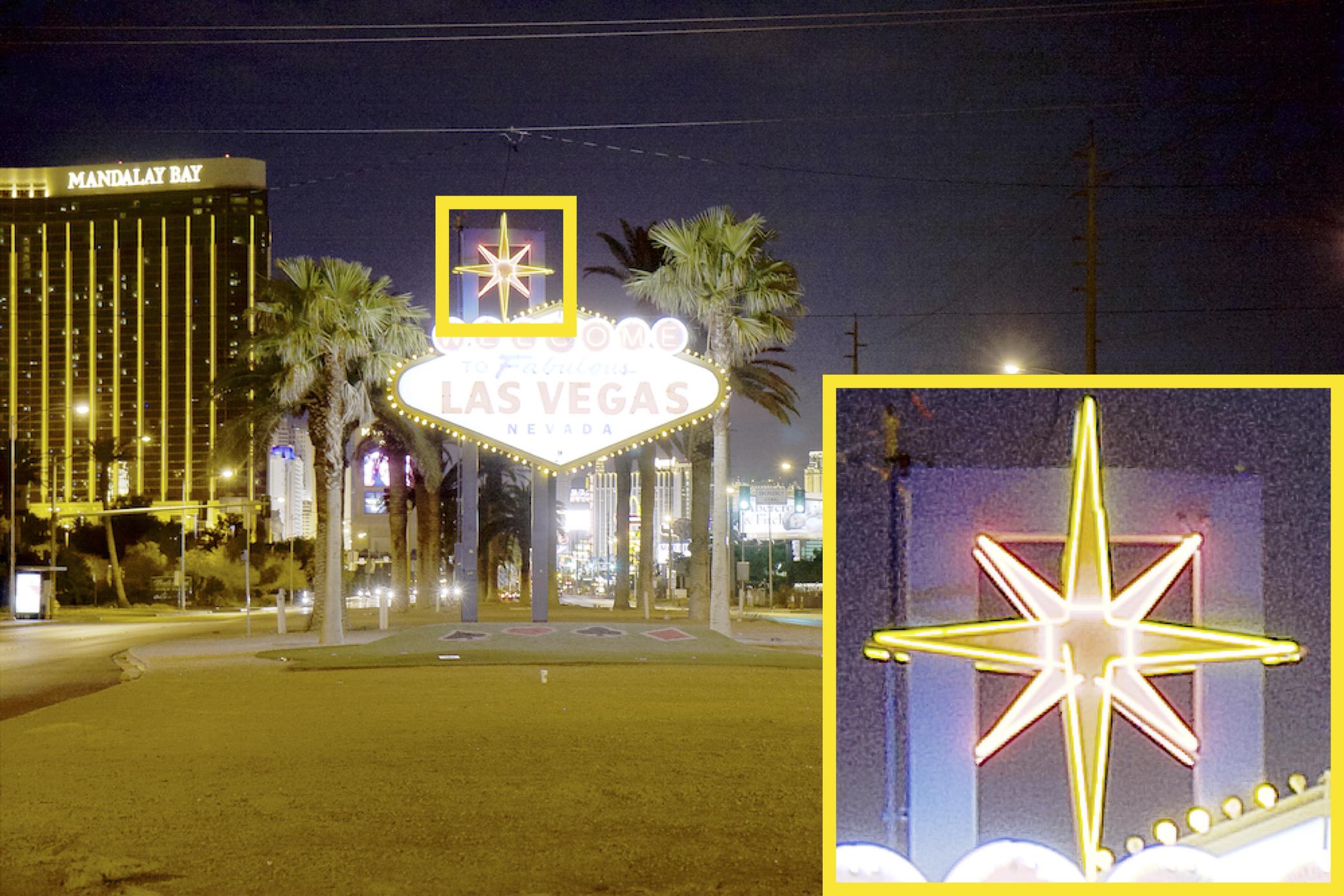}
        \caption{SCI \cite{ma2022toward}}
    \end{subfigure}%
    \begin{subfigure}[b]{0.5\textwidth}
        \centering
        \includegraphics[width=.7\linewidth]{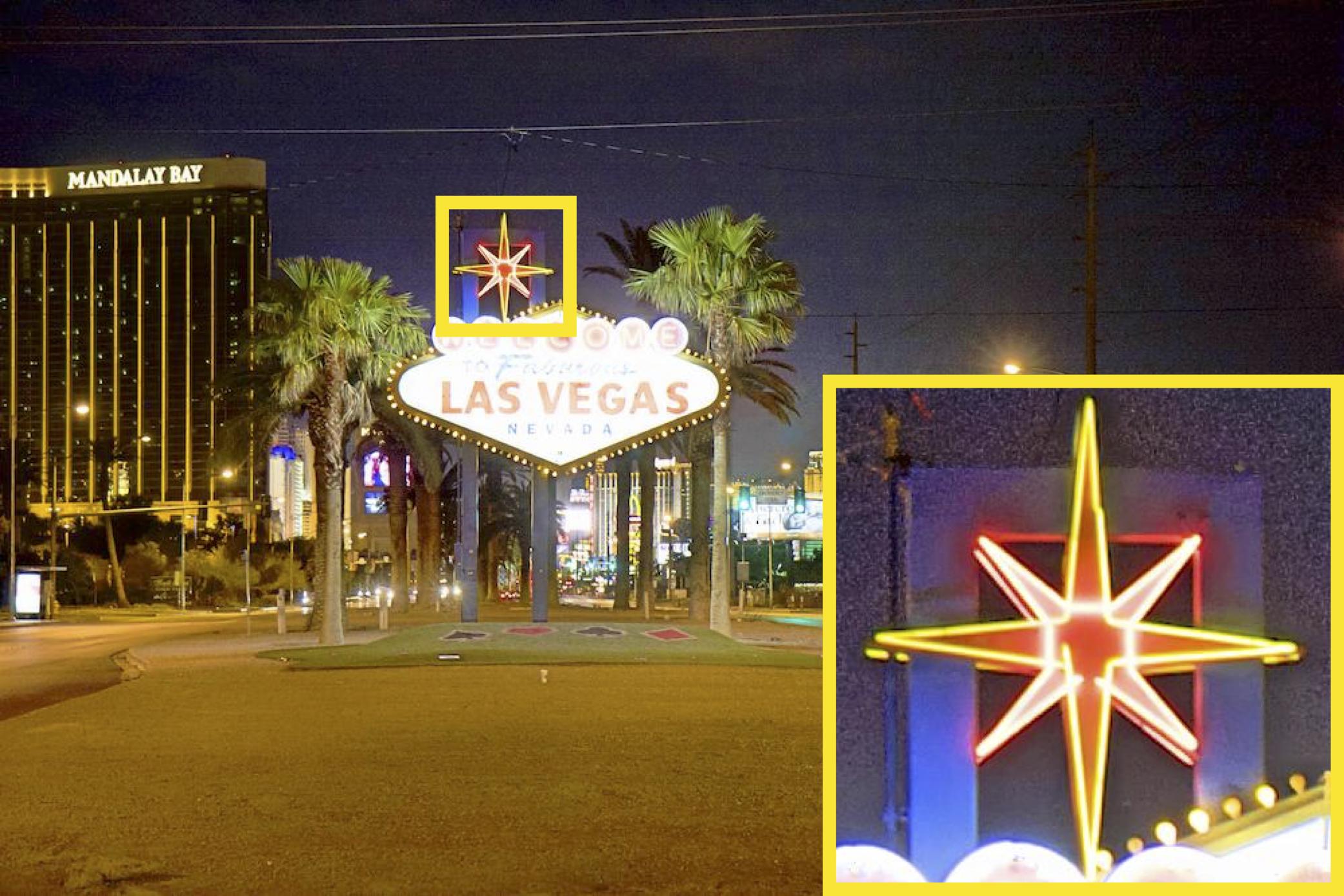}
        \caption{Ours}
    \end{subfigure}
    \caption{An example of a challenging real night image from the MIT dataset \cite{fivek} that illustrates the LLIE task. The intense light emitted by the sign results in significant under-exposure of other areas in the image. Our approach excels in recovering these regions with superior accuracy compared to state-of-the-art (SOTA) methods, which tend to overexpose well-lit areas and distort colors. We emphasize that the compared methods were pretrained on challenging low-light images, whereas our method restores the image in a zero-shot setting without any prior learning.}\label{fig:intro}
\end{figure}

Images captured under low-light conditions frequently contain substantial dark areas alongside regions that are overly exposed, resulting in the distortion of visual content. This degradation not only makes certain parts of the image imperceivable to human observers but also detrimentally impacts the efficacy of computer vision algorithms \cite{huang2020real, li2021photon, jia2023three, xu2021exploring}.

The task of Low-Light Image Enhancement (LLIE) has emerged as a prominent area of focus within the field. Besides early methods such as histogram equalization \cite{pizer1990contrast}, subsequent approaches based on the Retinex theory \cite{guo2016lime, ng2011total, ren2020lr3m, hao2020low} and convolutional neural network-based techniques (CNNs) for illumination estimation and deep low-light image representation have garnered considerable attention \cite{wu2022uretinex, zhang2021beyond, zhang2019kindling, lore2017llnet, yang2020fidelity}.

Deep learning algorithms mostly aim to achieve image quality through supervised learning to learn direct mapping between \textbf{paired} low-light and normal-light images \cite{chen2018learning, dong2022abandoning, kim2021representative, koh2022bnudc, zhang2022deep}, while unsupervised methods rely only on training with low-light images \cite{guo2020zero, jiang2021enlightengan, ma2022toward}. Nevertheless, convolutional processing is typically constrained by its inductive biases \cite{naseer2021intriguing}. Fully convolutional networks rely exclusively on local information, which hampers their ability to tackle large dark areas effectively. Therefore, CNN-based methods lack sensitivity in capturing precise representations required for accurate low-light image enhancement.

In this paper, we present a novel approach for enhancing low-light images by leveraging the neural implicit representation (NIR) of illumination, inferred directly from single input images. Unlike conventional methods that aim to learn direct mappings between RGB color spaces, our proposed method encodes spatial coordinates to generate corresponding image representations in HSV space. Inspired by recent advancements in implicit function modelling for tasks such as 3D shape reconstruction \cite{deng2020nasa, genova2020local, genova2019learning} and image processing \cite{chen2021learning, sitzmann2020implicit, yang2023implicit}, we use a network as an implicit function to map 2D image coordinates to the illumination field conditioned on original image values. The estimated illumination field is then used to enhance the low-light image according to principles of Retinex theory\cite{land1971lightness}. This innovative approach allows for zero-shot network training, removing the reliance on paired images or specific training sets of low-light images. Consequently, our approach offers enhanced flexibility, circumventing issues related to domain gaps and the need for retraining under varying under-exposure levels. 
Our method has the potential to significantly advance low-light image enhancement techniques, offering a promising avenue for real-world applications in computer vision. The contributions of this paper are as follows:
\begin{itemize}
    \item We present a novel formulation of the low-light image enhancement challenge by mapping 2D image coordinates to the illumination component using a network as the neural implicit function, conditioned on the original image intensity values in a zero-shot setting. This novel approach circumvents the need for prior training, thus mitigating domain gap concerns.
    
    \item We incorporate guided filtering into our framework, significantly reducing computational overhead for image restoration. This integration ensures that the runtime remains largely unaffected by increases in the input image size, thereby maintaining efficiency without compromising performance.

    \item We conducted extensive evaluations of our method across three diverse datasets containing natural images as well as microscopy images. The consistent and superior performance across these varied datasets underscores the robustness and versatility of our approach, showcasing its effectiveness in enhancing low-light images in both general and specialized imaging contexts.
\end{itemize}

\section{Related Work}
\label{sec:related-work}

\subsubsection{Classic Methods for LLIE.} Conventional LLIE methods often rely on the Retinex theory \cite{land1971lightness}, which formulates image formation as a multiplication of two key elements: illumination and reflectance (Equation \ref{eq:retinex}). Some approaches adopt a variational framework to estimate reflectance and illumination, and subsequently refining illumination to enhance the input image \cite{fu2016weighted, kimmel2003variational, li2021photon}. Other methods utilize the Maximum-a-Posteriori (MAP) framework to restore low-light images by imposing priors on reflectance and illumination components \cite{guo2016lime, hao2020low, li2018structure}, such as total variation or structure-aware regularization. However, the drawback of these methods is their reliance on assumed priors, controlled by numerous parameters. Consequently, this can lead to distorted colors or inadequate balance between heavily over- and under-exposed regions, limiting their practical applicability in real-world scenarios.

\subsubsection{Network-based Methods for LLIE.} The advancement of LLIE has been significantly driven by the use of CNNs \cite{li2021low}, improving the restoration accuracy. While some approaches incorporate the Retinex theory into their models \cite{wu2022uretinex, Chen2018Retinex, zhang2021beyond, zhang2019kindling}, others opt for directly predicting enhanced images from low-light inputs through supervised learning strategies \cite{chen2018learning, lore2017llnet, dong2022abandoning, kim2021representative}. However, the required paired training data for these supervised approaches are often difficult to obtain in practice \cite{ma2022toward}. Recent methods have sought to train restoration models exclusively on low-light images to address this limitation. For instance, ZeroDCE \cite{guo2020zero} utilizes a deep network to establish image-specific curve estimations. Liu \textit{et al.} \cite{liu2021retinex} introduced a Retinex-based unrolling framework employing architecture search, while Ma \textit{et al.} \cite{ma2022toward} proposed a self-calibrating module for image brightening through cascaded illumination learning. Additionally, Fu \textit{et al.} \cite{fu2023learning} presented an approach leveraging adaptive priors learned from pairs of low-light images to assist the Retinex decomposition. Despite being framed as zero-reference and self-supervised methods, they still depend on prior training on low-light images, making them less stable and challenging to use in real-world scenarios where the level of under-exposure may not align with that of the training data. Furthermore, the CNNs in these methods are highly computationally demanding, especially when dealing with high-resolution images.

\subsubsection{Neural Implicit Representations.} Although neural implicit representations (NIR) haven been adapted for a variety of image processing tasks, such as super-resolution \cite{chen2021learning, lee2022local}, image warping into continuous shapes \cite{lee2022learning}, tomographic reconstruction \cite{osti_1883143}, their application in low-light image enhancement (LLIE) is relatively unexplored. Closest to our proposed approach is NeRCo introduced by Yang \textit{et al.} \cite{yang2023implicit}, which uses a neural representation module to normalize the degradation condition of the input low-light image before enhancing it with a ResNet-based module. Our proposed approach significantly diverges from NeRCo, which uses NIR merely as a pre-processing step, relying on conventional CNNs for the enhancement. In contrast, our method leverages NIR directly to extract the illumination component, utilizing it to enhance the low-light image according to the Retinex theory.

\section{Method}
\label{sec:method}

\begin{figure}[t]
    \begin{subfigure}[b]{0.333\textwidth}
        \centering
        \includegraphics[width=.9\linewidth]{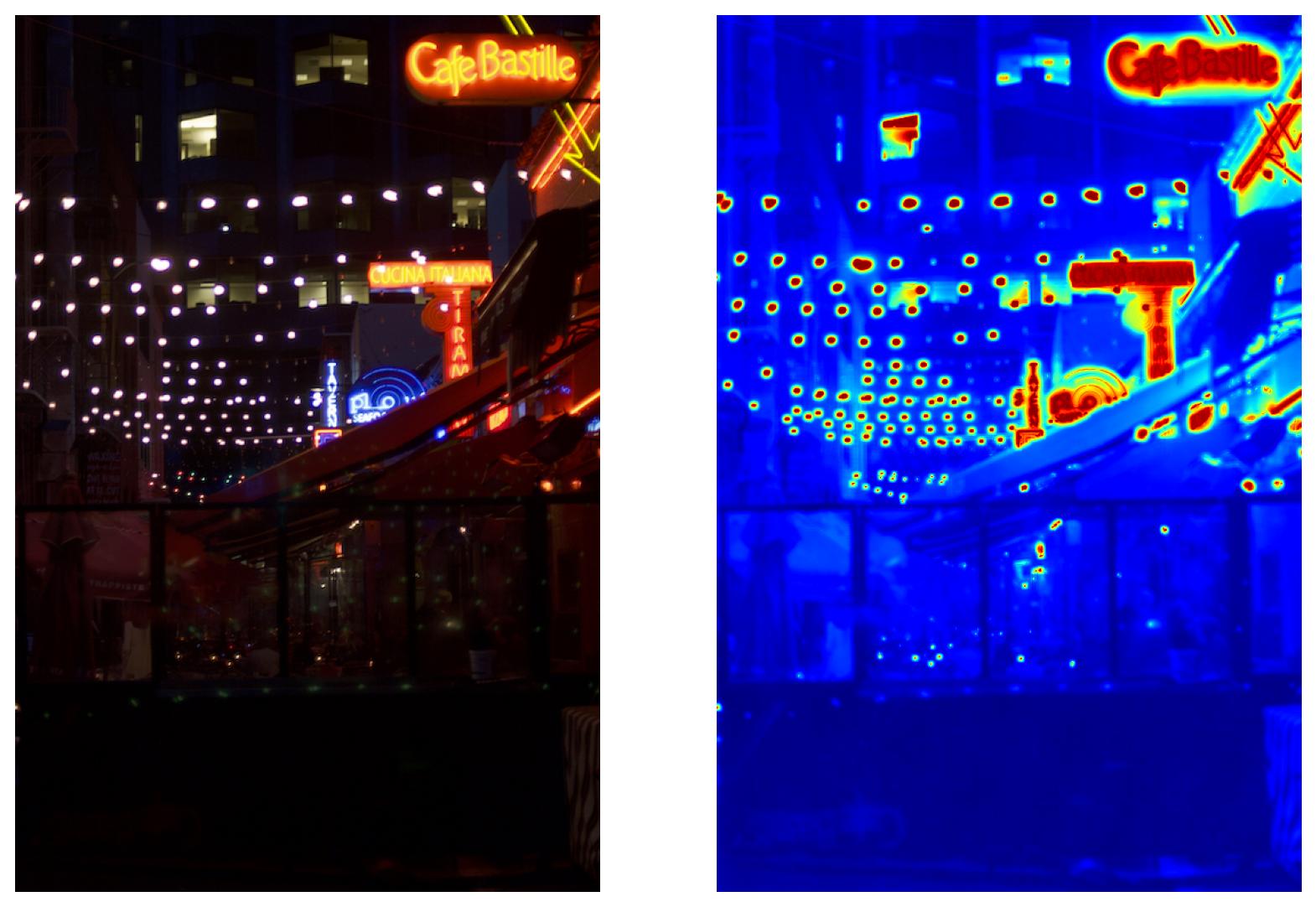}
        \caption{}
    \end{subfigure}%
    \hfill
    \begin{subfigure}[b]{0.333\textwidth}
        \centering
        \includegraphics[width=.9\linewidth]{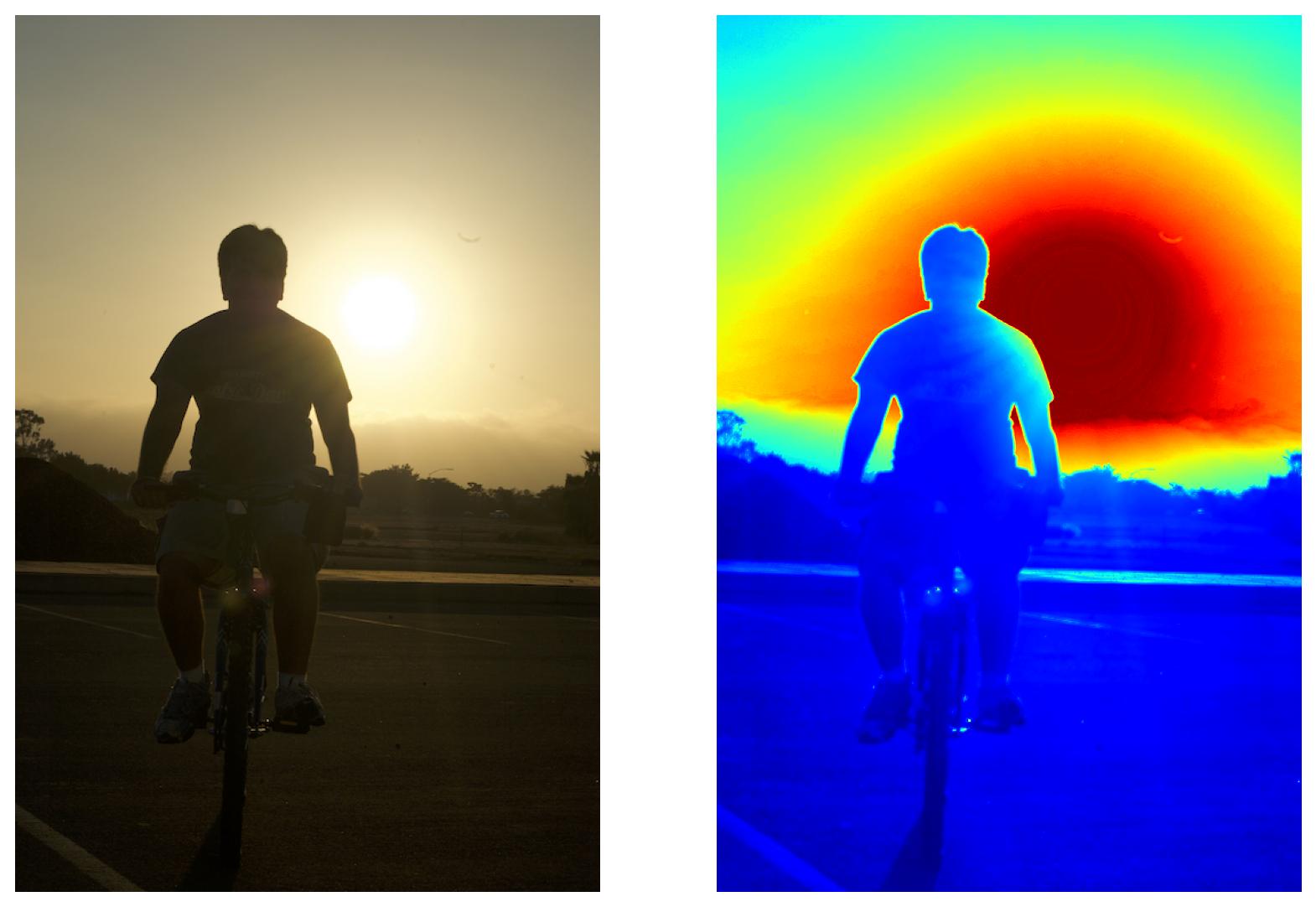}
        \caption{}
    \end{subfigure}%
    \hfill
    \begin{subfigure}[b]{0.333\textwidth}
        \centering
        \includegraphics[width=.9\linewidth]{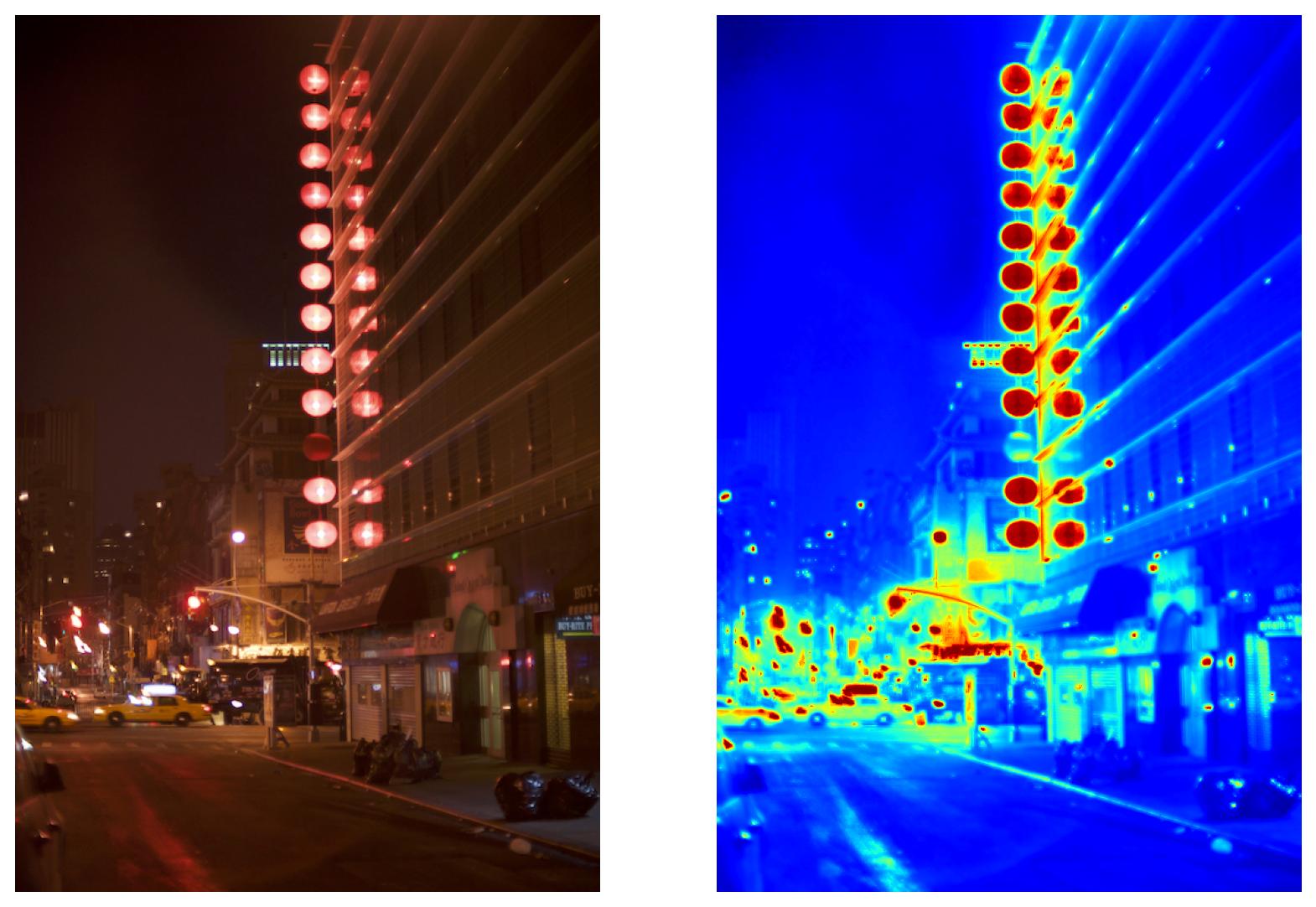}
        \caption{}
    \end{subfigure}%
    \caption{Instances of inferred illumination components represented by heatmaps based on images within the MIT dataset \cite{fivek}. These estimated illumination components are subsequently used to improve low-light images according to the principles of the Retinex theory. Leveraging this approach, the enhancement process aims to faithfully reproduce the perceived visual richness and detail even in challenging lighting conditions.}\label{fig:illu}
\end{figure}

\subsection{Illumination Learning}
\label{subsec:illu-learning}

The Retinex theory \cite{land1971lightness} establishes a foundational connection between the visual observation of images captured under low-light conditions, denoted as $\mathbf{y}$, and their corresponding high-quality representations under ideal lighting conditions, represented as $\mathbf{z}$. This connection is tied to the concept of an illumination component, denoted by $\mathbf{x}$, which indicates local light intensities in the image, thus yielding the observed low-light image as follows,
\begin{equation}
    \label{eq:retinex}
    \mathbf{y} = \mathbf{x} \odot \mathbf{z}.
\end{equation}

Within the domain of low-light image enhancement, the illumination component $\mathbf{x}$ assumes a central role, wielding significant influence over the overall perceived visual quality, and therefore the accurate estimation and manipulation of this illumination component is crucial to effectively enhance low-light images while simultaneously preserving essential details and mitigating the presence of undesirable artifacts.

We introduce a mapping $\mathcal{M}_\theta$ with parameters $\theta$ that aims to learn the illumination component to enhance the low-light image as follows,
\begin{equation}
  \mathcal{M}_\theta(\mathbf{y}) :
    \begin{cases}
      & \hat{\mathbf{x}}=\mathbf{y} + f_\theta(\cdot),\\
      & \hat{\mathbf{z}}=\mathbf{y}\oslash\hat{\mathbf{x}}
    \end{cases}.
\end{equation}
This process is inspired by the consensus that the illumination and low-light observation are similar, and there exists a linear connection in most areas. By learning the residual with function $f_\theta(\cdot)$, we reduce the computational difficulty and improve the stability and robustness for exposure control. 

\begin{figure}[t]
    \centering
    \begin{subfigure}[b]{0.5\textwidth}
        \centering
        \includegraphics[width=.7\linewidth]{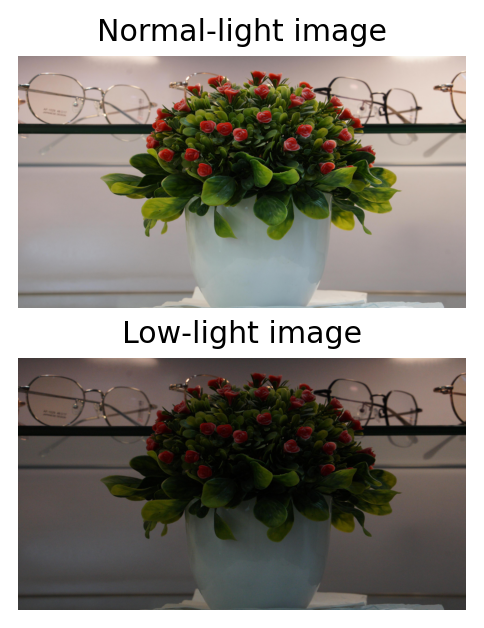}
        \caption{}
    \end{subfigure}%
    \begin{subfigure}[b]{0.5\textwidth}
        \centering
        \includegraphics[width=.95\linewidth]{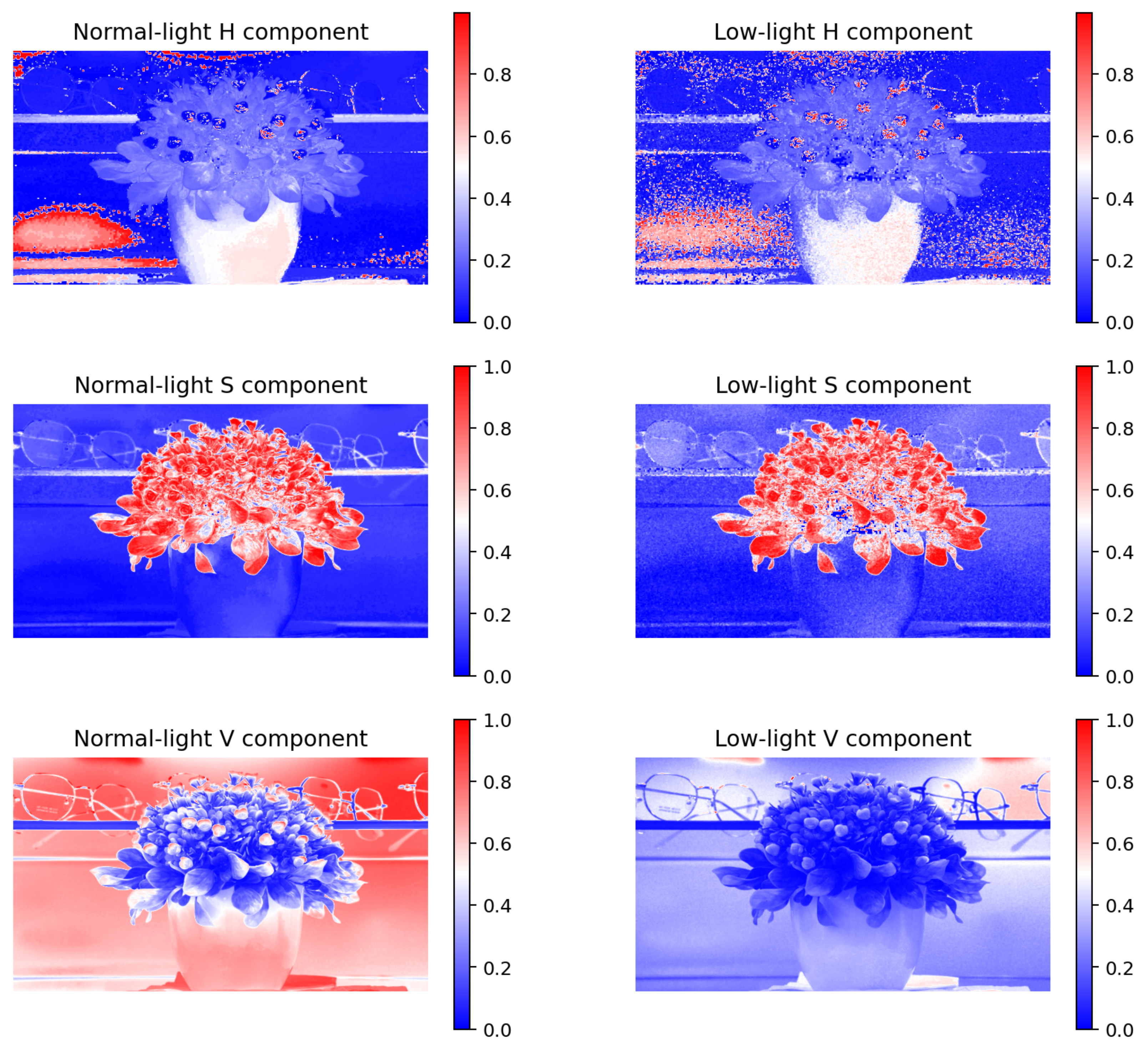}
        \caption{}
    \end{subfigure}%
    \caption{The motivation behind the design decisions of our method stems from a comparative analysis of the HSV components of normal-light and low-light images from the UHD-LL dataset \cite{Li2023ICLR}. This overview shows that, while the Hue and Saturation components display only minimal differences, the primary distinction between the two images lies in the Value component.}\label{fig:hsv}
\end{figure}

Unlike the recent work \cite{guo2020zero, liu2021retinex, ma2022toward, yang2023implicit}, we deliberately departure from RGB image representations $\mathbf{y}_C\in\mathbb{R}^{H\times W}$ where $C\in\{R,G,B\}$ in favor of the Hue-Saturation-Value (HSV) cylindrical coordinate system, $\mathbf{y}_{\overline{C}}\in\mathbb{R}^{H\times W}$ where ${\overline{C}}\in\{H,S,V\}$. This choice is motivated by the inherent advantages of HSV in disentangling color and brightness components, and thereby allowing for improved manipulation capabilities within low-light representations. A plausible conjecture within our framework is that the Hue and Saturation components exhibit a degree of similarity between corresponding regions within both the low-light and enhanced images, as seen in Fig. \ref{fig:hsv}. Therefore, our proposed mapping is done directly on the Value component (brightness) of the HSV image representation,
\begin{equation}
  \mathcal{M}_\theta(\mathbf{y}_V) :
    \begin{cases}
      & \hat{\mathbf{x}}_V=\mathbf{y}_V + f_\theta(\cdot),\\
      & \hat{\mathbf{z}}_V=\mathbf{y}_V\oslash\hat{\mathbf{x}}_V
    \end{cases}.
\end{equation}
The culmination of our image enhancement process involves the combination of the original Hue and Saturation components alongside the enhanced Value component, resulting in composite representation $\hat{\mathbf{z}}\in\{\mathbf{y}_H,\mathbf{y}_S,\hat{\mathbf{z}}_V\}$, which integrates the color attributes extracted from the input image and refined brightness. 

\subsection{Problem Formulation With Implicit Function}
\label{subsec:NIF}

We construct a model, represented by $f_\theta$, aimed at extracting the illumination component $\mathbf{x}$ by employing a parameterized MLP to embed the input image. Differing from prior approaches that establish mappings between color spaces (e.g., from RGB to RGB), we reframe the problem as a mapping from 2D coordinates to the elements of the Value component, conditioned on the original Value component,
\begin{equation}
    f:(p,\mathcal{N}(p))\rightarrow r,
\end{equation}
where $p_i=(a_i,b_i)$ are the coordinates of the Value component elements, and $r_i$ is the corresponding pixel intensity. The conditioning is done based on a window of elements of the original Value component with a size $W\times W$ centered around the pixel indexed by $p_i$, specifically,
\begin{equation}
    f_\theta:(p_i,\mathcal{N}(p_i))\rightarrow r_i.
\end{equation}
We depart from conventional neural implicit representation functions by introducing a novel incorporation of local context information as input. This is motivated by the intention to provide richer information about the input data, thus enhancing the quality of output representations. By integrating local context alongside the coordinates, our model gains a more comprehensive understanding of the spatial relationships and contextual connections within the scene. This strategy enables our network to capture intricate details and nuanced variations, resulting in more informative and robust output representations. 

Similarly to \cite{li2023metadata}, we build the MLP structure with SIREN layers \cite{sitzmann2020implicit}, where the pixel coordinates $p_i$ and the context-window values $\mathcal{N}(p_i)$ are separated into two branches and then concatenated into a single output branch. We use $256$ channels for the hidden layers and the output of the two input branches is reduced in half before being concatenated and passed to the output layers.

\begin{figure}[t]
    \centering
    \includegraphics[width=\linewidth]{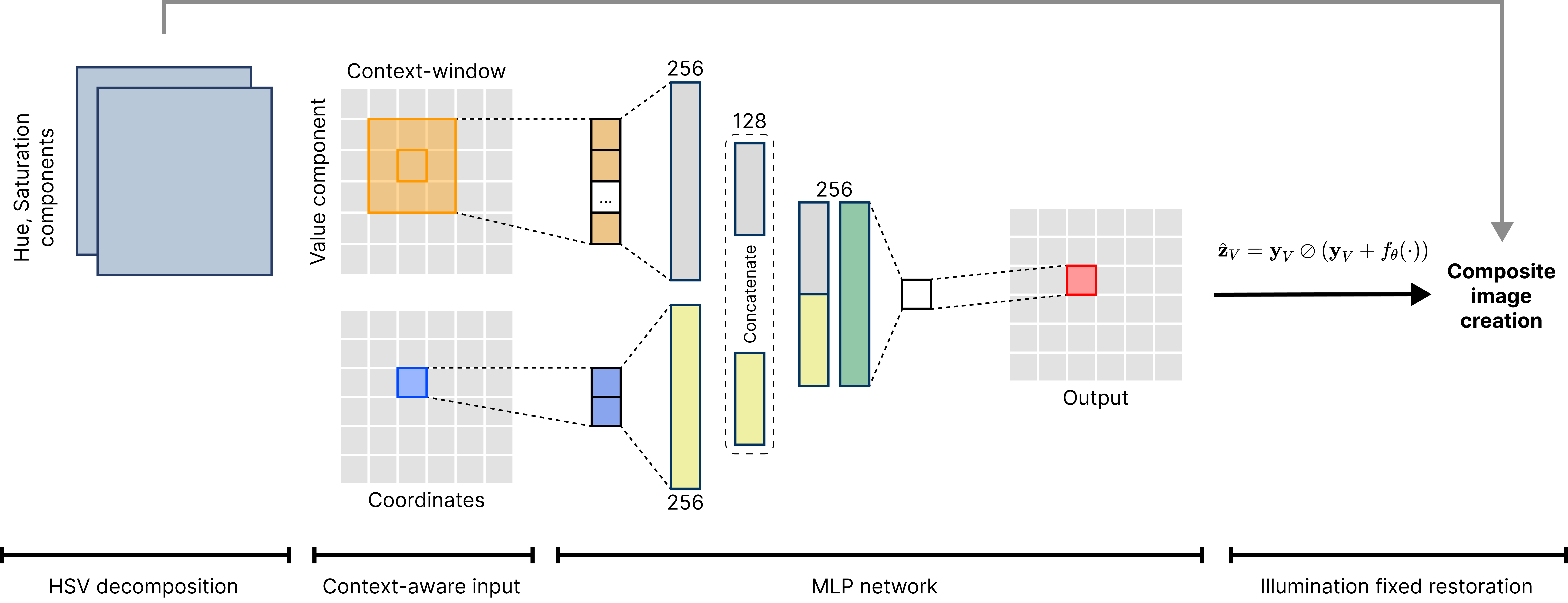}
    \caption{Our proposed framework begins with the extraction of the Value component from the HSV image representation. Subsequently, we employ a NIR model to infer the illumination component which is an essential part for effective enhancement of the input low-light image. This refined Value component is then reintegrated with the original Hue and Saturation components, forming a comprehensive representation of the enhanced image. The architecture of CoLIE involves dividing the inputs into two distinct parts: the elements of the Value component and the coordinates of the image. Each of these components is subject for regularization with unique parameters within their respective branches. By adopting this structured approach, our framework ensures precise control over the enhancement process.}
    \label{fig:architecture}
\end{figure}

\subsection{Zero-Shot Training}
\label{subsec:loss}

Considering the inaccuracy of existing paired training data and the necessity of model retraining for each image domain and degree of under-exposure, we propose a no-reference learning to mitigate those concerns. We define the total loss as 
\begin{equation}
    \mathcal{L}_\textit{total}=\alpha\mathcal{L}_\textit{f} + \beta\mathcal{L}_\textit{s} + \gamma\mathcal{L}_\textit{exp} +
    \delta\mathcal{L}_\textit{spa},
\end{equation}
where $\mathcal{L}_\textit{f},\mathcal{L}_\textit{s},\mathcal{L}_\textit{exp}$ and $\mathcal{L}_\textit{spa}$ represent fidelity, smoothness, exposure and sparsity loss, respectively, and $\alpha,\beta,\gamma,\delta$ are loss balancing parameters.

We use the mean squared error (MSE) for the fidelity loss $
\mathcal{L}_f$ to maintain pixel-level consistency between the estimated illuminates component and the low-light observation. The smoothness, which is one of the defining factors of the illumination field according to the Retinex theory, and is a broad consensus in this task, is enforced by the total variation (TV) as follows,
\begin{equation}
    \mathcal{L}_\textit{s}=\left(\lVert\nabla_i \hat{\mathbf{x}}_V\rVert_2 + \lVert\nabla_j \hat{\mathbf{x}}_V\rVert_2\right)^2,
\end{equation}
where $\nabla_i$ and $\nabla_j$ represent the vertical and horizontal gradient operations, respectively.

To restrain issues stemming from under- and over-exposed areas, we regulate the intensity levels within the illumination field. We control the distance between the gamma-controlled region and the optimally-intense threshold denoted as $L$. In our experimental setting, unless explicitly stated otherwise, we set $L$ as $0.5$ to achieve most visually pleasing results based on our empirical evidence (see Section \ref{sec:ablation}). The loss function $\mathcal{L}_\textit{exp}$ can be expressed as follows,
\begin{equation}
    \mathcal{L}_\textit{exp}=\frac{1}{N}\sum_{k=1}^N\left\lVert\sqrt{\mathcal{T}_k} - L\right\rVert_2,
\end{equation}
where $N$ represents the number of non-overlapping local regions, and $\mathcal{T}$ denotes the average intensity value of a given local region within the illumination field.

In order to preserve the fidelity of low-light areas and mitigate the risk of over-exposure in dim regions, we introduce a sparsity term into the loss function. This sparsity loss penalizes high-intensity values, thus promoting a more balanced illumination across the image.
\begin{equation}
    \mathcal{L}_\textit{spa}=\frac{1}{M}\sum_{l=1}^M \left|\left(\hat{\mathbf{z}}_V\right)_l\right|,
\end{equation}
where $M$ denotes the total number of pixels in the image. By incorporating the sparsity loss term, we encourage the preservation of dark regions while discouraging the occurrence of excessively bright areas, thereby enhancing the overall fidelity of the reconstructed image.

To address the computational demands inherent in handling (ultra) high-definition images, we process the image in its low-resolution representation $\mathbf{y}^\textit{LR}$ to obtain $\hat{\mathbf{z}}^\textit{LR}$, and then adopt the guided filter \cite{wu2018fast} to restore the original resolution. This strategic utilization of guided filtering not only mitigates computational costs and scalability challenges but also ensures the preservation of image fidelity throughout the upscaling process.

\section{Experiments}

\subsubsection{Implementation Details.} During training, we use the Adam optimizer \cite{kingma2014adam} with parameters $\beta_1=0.9$ and $\beta_2=0.999$. The learning rate was set to $10^{-5}$ to initialize the training for 100 epochs. The regularisation parameters were set as $\alpha=1, \beta=20, \gamma=8, \delta=5$, and the size of the context window was set to $7\times 7$. Both input branches for the context window and coordinates comprised two layers each, with their outputs passed to two separate output layers for the final prediction. Prior to training, the input images were downscaled to the resolution of $256 \times 256$ pixels to enhance processing speed. After training, the outputs were upscaled using the guided filter \cite{wu2018fast} to restore the original dimensions of the image. To utilize the context window even for pixels located at the border of the image, we employ reflection padding to extend the image.

\subsubsection{Compared Methods.} We include 10 state-of-the-art (SOTA) methods with different pretrained weights (16 models in total) in our benchmarking study and performance evaluation. The methods cover various approaches, including supervised methods such as DeepLPF \cite{fan2022half}, HWMNet \cite{fan2022half}, RetinexNet \cite{Chen2018Retinex}, Zhao \textit{et al.} \cite{Zhao_2021_ICCV}, unsupervised methods like ZeroDCE \cite{guo2020zero}, EnGAN \cite{jiang2021enlightengan}, RUAS \cite{liu2021retinex}, SCI \cite{ma2022toward}, PairLIE \cite{fu2023learning}, and one optimization-based method, LIME \cite{guo2016lime}. The methods have been trained with diverse paired and unpaired low-light images including the MIT \cite{fivek}, LOL \cite{Chen2018Retinex}, LOL-v2 \cite{Yang_2020_CVPR}, SICE \cite{Cai2018deep}, LSRW \cite{hai2023r2rnet} and DarkFace \cite{yang2020advancing} datasets.

For quantitative benchmarking with two full-reference metrics, SSIM and PSNR, we use the testing set of the UHD-LL dataset \cite{Li2023ICLR} and randomly select 200 images from the MIT \cite{fivek} dataset. For qualitative analysis and downstream task evaluation, we use the DarkFace dataset \cite{yang2020advancing}.

\subsubsection{Performance Evaluation.} We conduct a reference-based quantitative analysis between our approach and the SOTA on the UHD-LL and MIT datasets, as presented in Tables \ref{table:uhd} and \ref{table:mit}, respectively. Our method demonstrates competitive performance, achieving the highest PSNR score on the UHD-LL dataset. Furthermore, our method surpasses the performance of the compared methods on the MIT dataset, showcasing its effectiveness in enhancing low-light images. 

\begin{figure}[t]
    \begin{subfigure}[b]{0.33\textwidth}
        \centering
        \includegraphics[width=.9\linewidth]{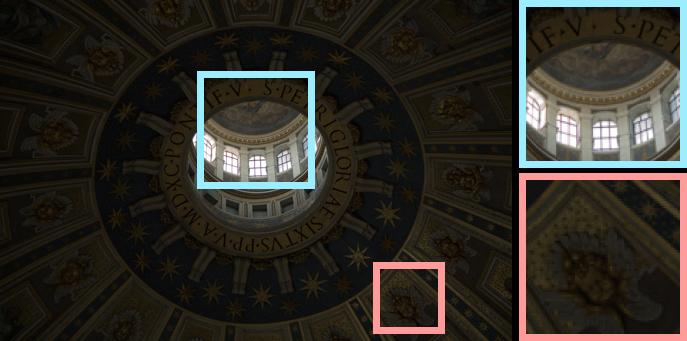}
        \caption{Input}
    \end{subfigure}%
    \begin{subfigure}[b]{0.33\textwidth}
        \centering
        \includegraphics[width=.9\linewidth]{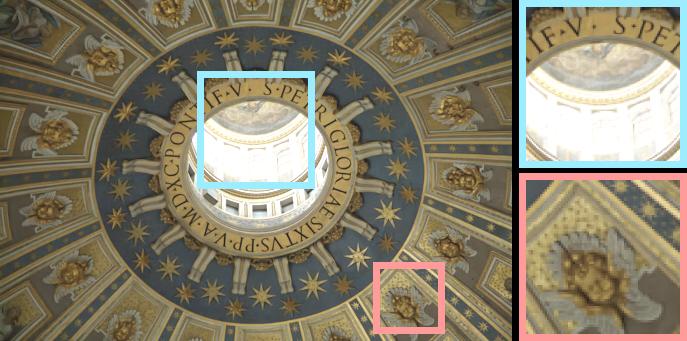}
        \caption{HWMNet \cite{fan2022half}}
    \end{subfigure}%
    \begin{subfigure}[b]{0.33\textwidth}
        \centering
        \includegraphics[width=.9\linewidth]{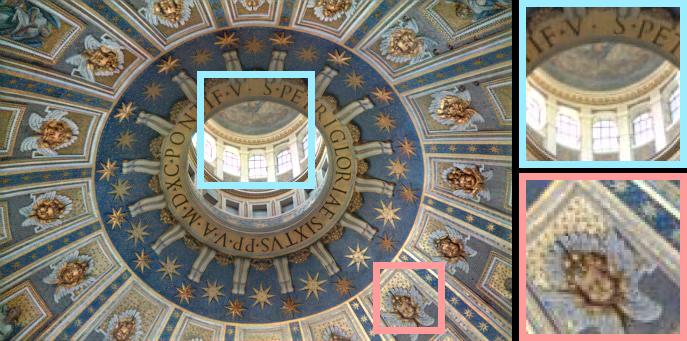}
        \caption{EnGAN \cite{jiang2021enlightengan}}
    \end{subfigure}%
    \newline
    \begin{subfigure}[b]{0.33\textwidth}
        \centering
        \includegraphics[width=.9\linewidth]{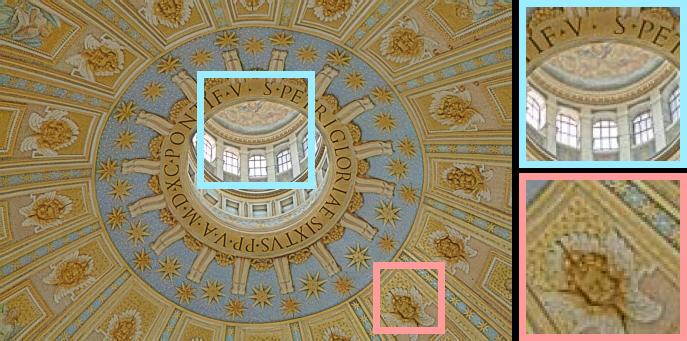}
        \caption{RetinexNet \cite{Chen2018Retinex}}
    \end{subfigure}%
    \begin{subfigure}[b]{0.33\textwidth}
        \centering
        \includegraphics[width=.9\linewidth]{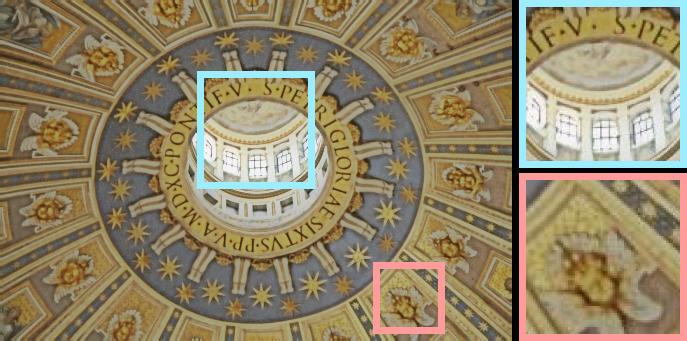}
        \caption{PairLIE \cite{fu2023learning}}
    \end{subfigure}%
    \begin{subfigure}[b]{0.33\textwidth}
        \centering
        \includegraphics[width=.9\linewidth]{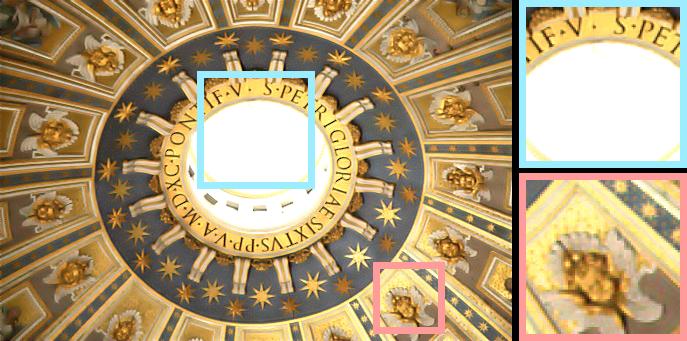}
        \caption{RUAS-LOL \cite{liu2021retinex}}
    \end{subfigure}%
    \newline
    \begin{subfigure}[b]{0.33\textwidth}
        \centering
        \includegraphics[width=.9\linewidth]{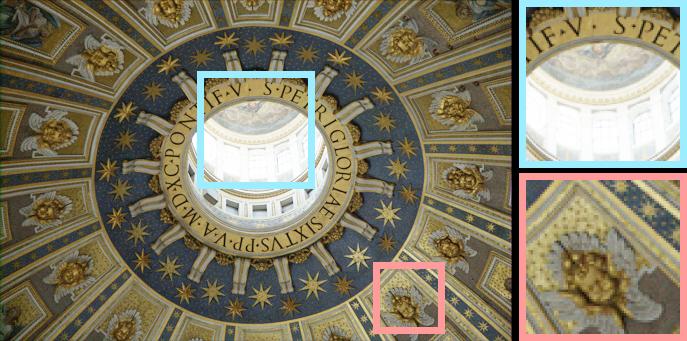}
        \caption{SCI-medium \cite{ma2022toward}}
    \end{subfigure}%
    \begin{subfigure}[b]{0.33\textwidth}
        \centering
        \includegraphics[width=.9\linewidth]{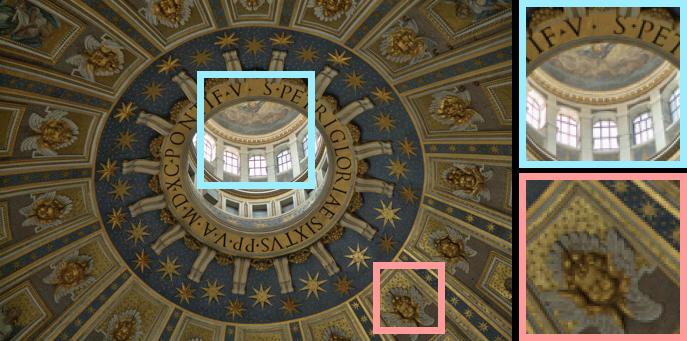}
        \caption{Ours}
    \end{subfigure}%
    \begin{subfigure}[b]{0.33\textwidth}
        \centering
        \includegraphics[width=.9\linewidth]{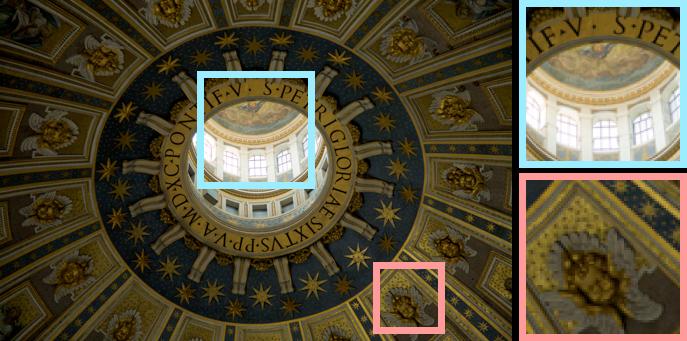}
        \caption{GT}
    \end{subfigure}%
    \caption{Visual quality comparison with SOTA methods on a real-world low-light image from the MIT dataset. Our method does not over-expose well-lit regions and preserves the original colors.}\label{fig:mit-qualitative}
\end{figure}

\begin{figure}[t]
    \begin{subfigure}[b]{0.33\textwidth}
        \centering
        \includegraphics[width=.9\linewidth]{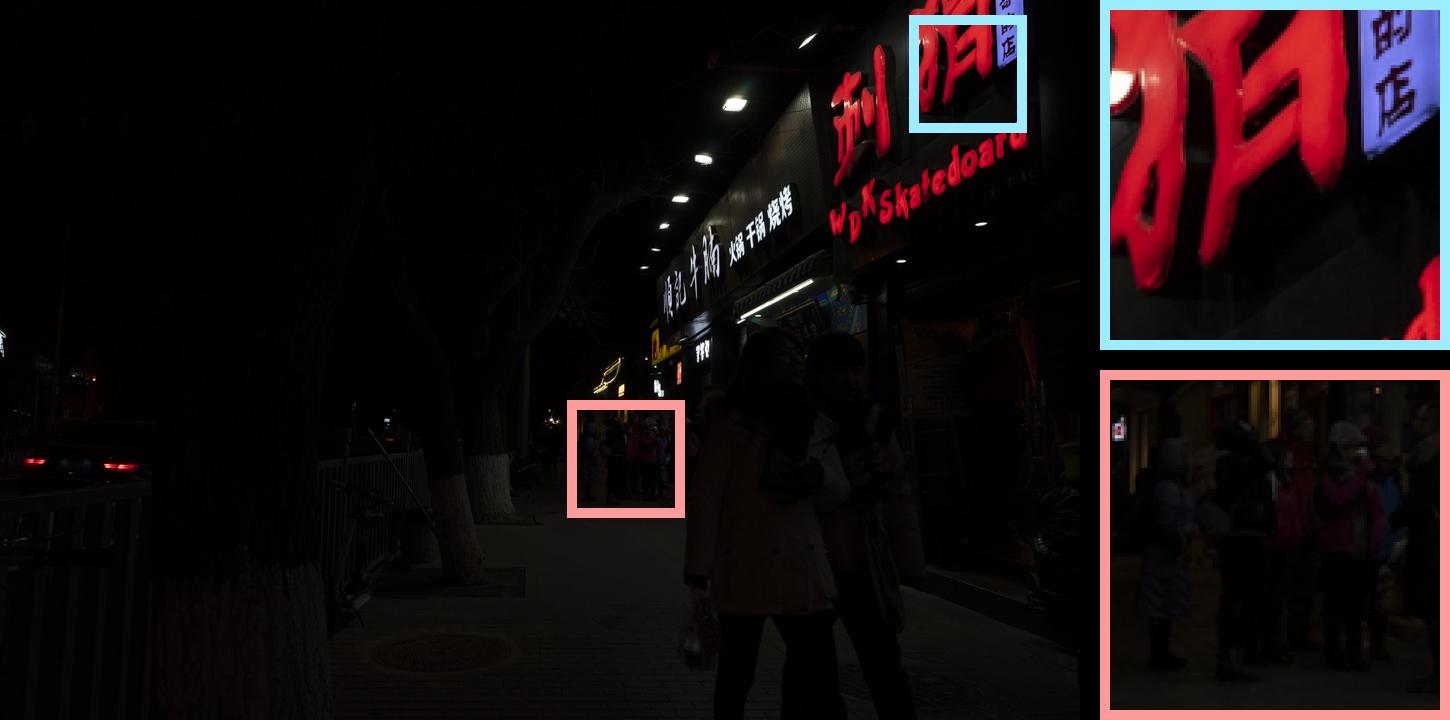}
        \caption{Input}
    \end{subfigure}%
    \begin{subfigure}[b]{0.33\textwidth}
        \centering
        \includegraphics[width=.9\linewidth]{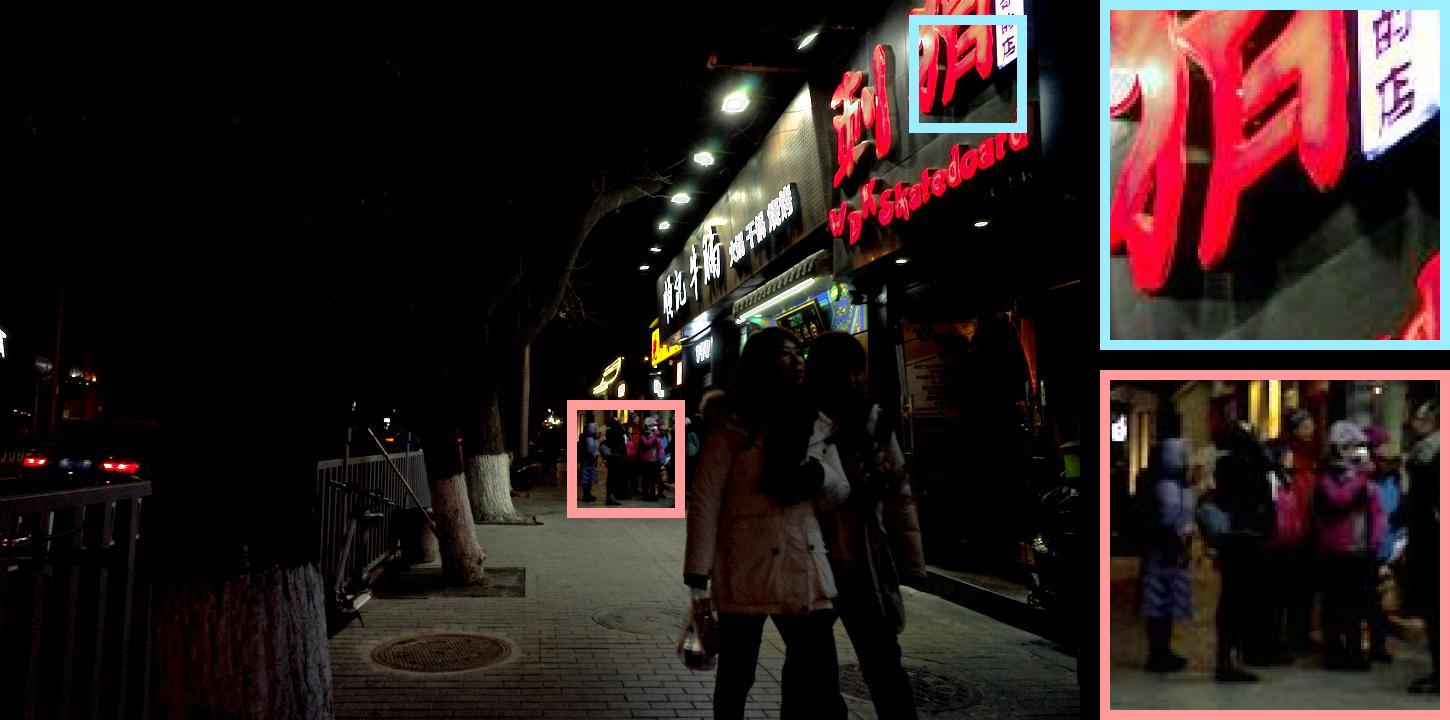}
        \caption{DeepLPF$^\dagger$ \cite{Moran_2020_CVPR}}
    \end{subfigure}%
    \begin{subfigure}[b]{0.33\textwidth}
        \centering
        \includegraphics[width=.9\linewidth]{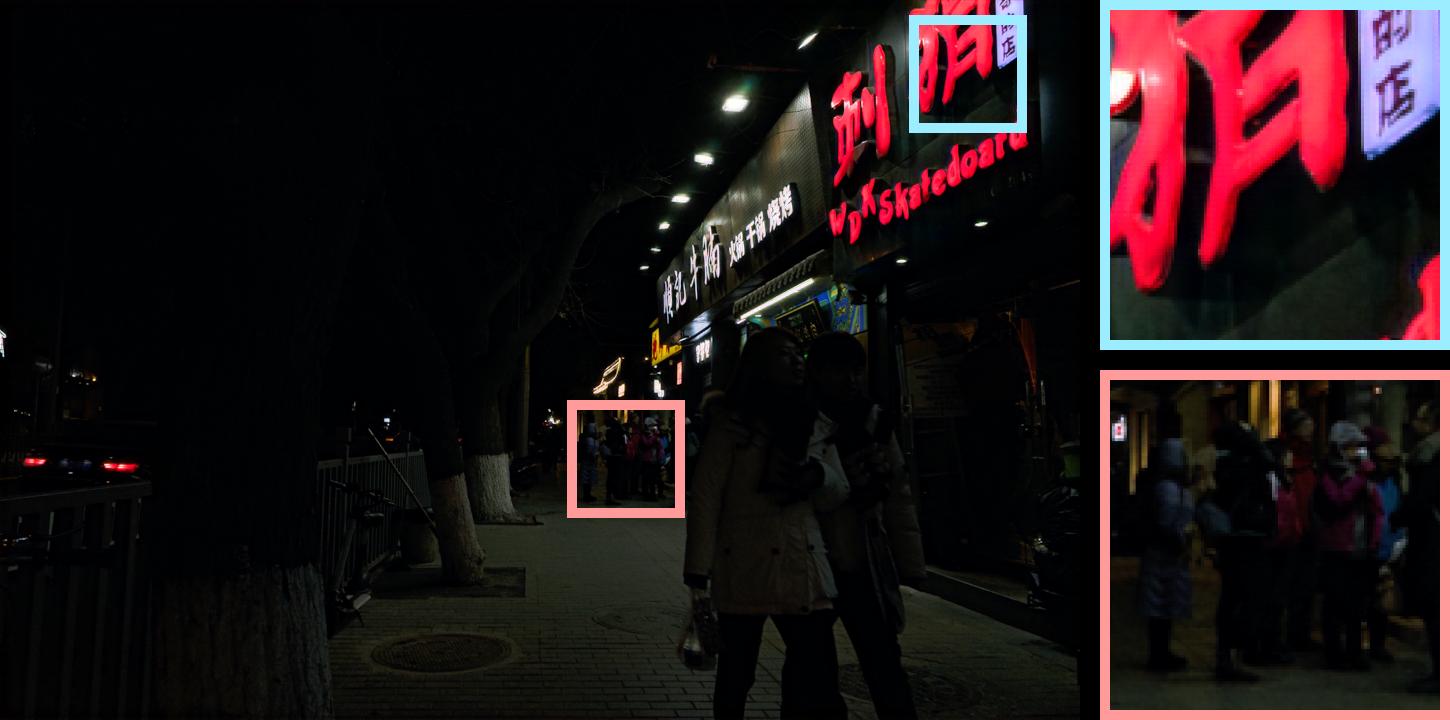}
        \caption{HWMNet$^\dagger$ \cite{fan2022half}}
    \end{subfigure}%
    \newline
    \begin{subfigure}[b]{0.33\textwidth}
        \centering
        \includegraphics[width=.9\linewidth]{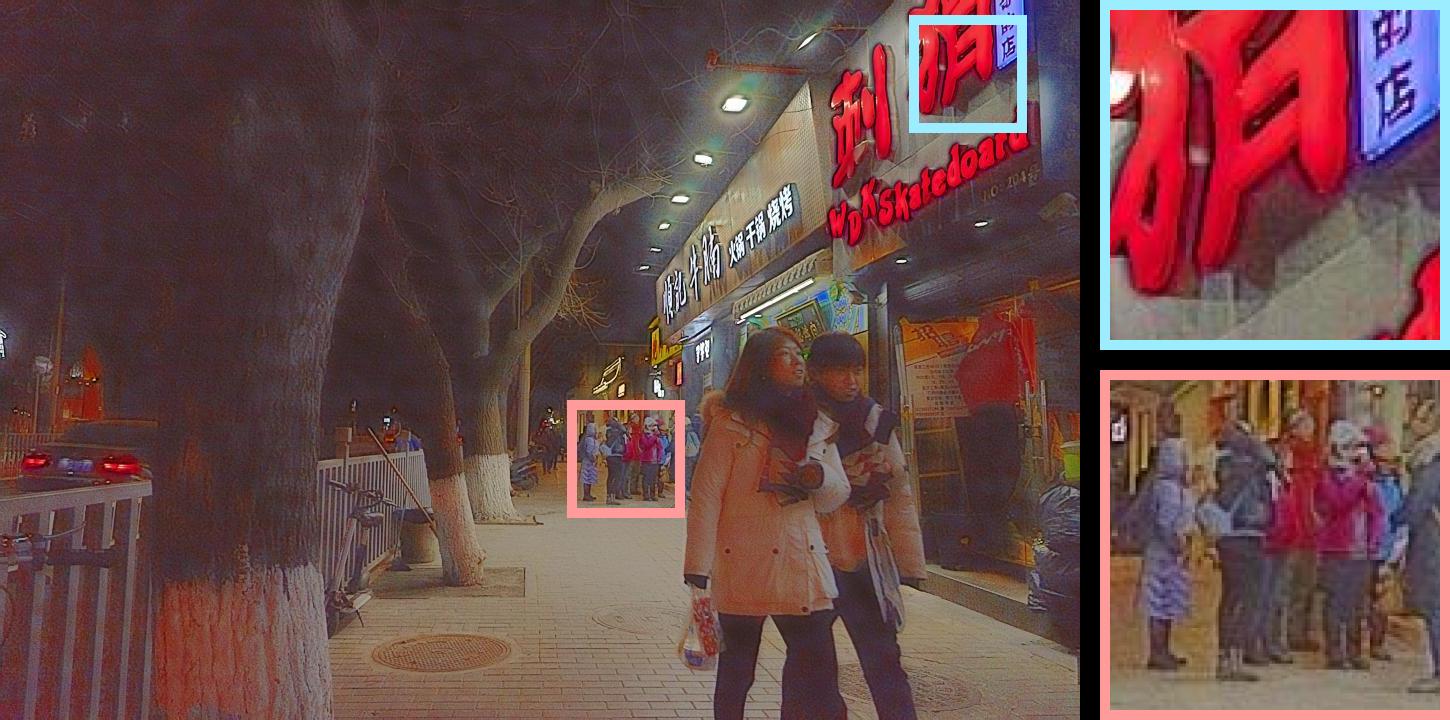}
        \caption{RetinexNet$^\dagger$ \cite{Chen2018Retinex}}
    \end{subfigure}%
    \hfill
    \begin{subfigure}[b]{0.33\textwidth}
        \centering
        \includegraphics[width=.9\linewidth]{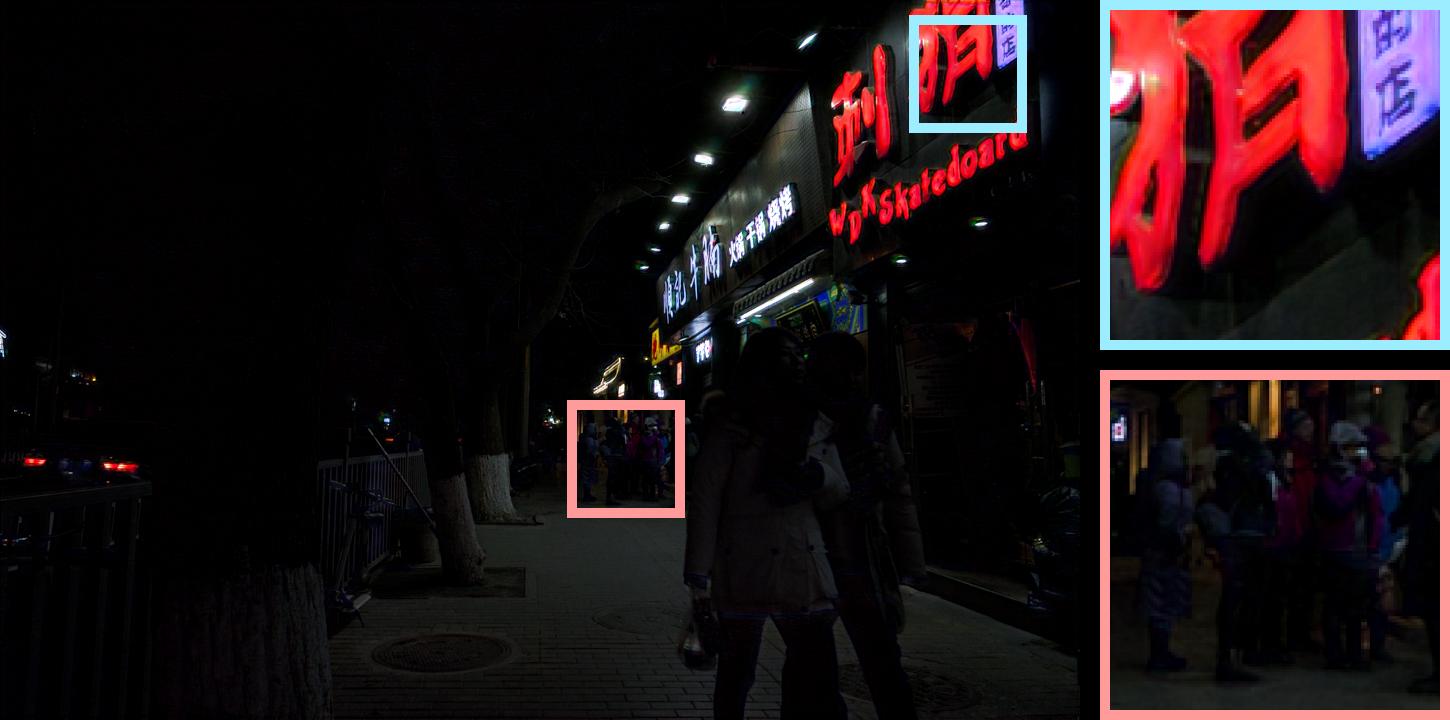}
        \caption{Zhao \textit{et al.}$^\dagger$ \cite{Zhao_2021_ICCV}}
    \end{subfigure}%
    \hfill
    \begin{subfigure}[b]{0.33\textwidth}
        \centering
        \includegraphics[width=.9\linewidth]{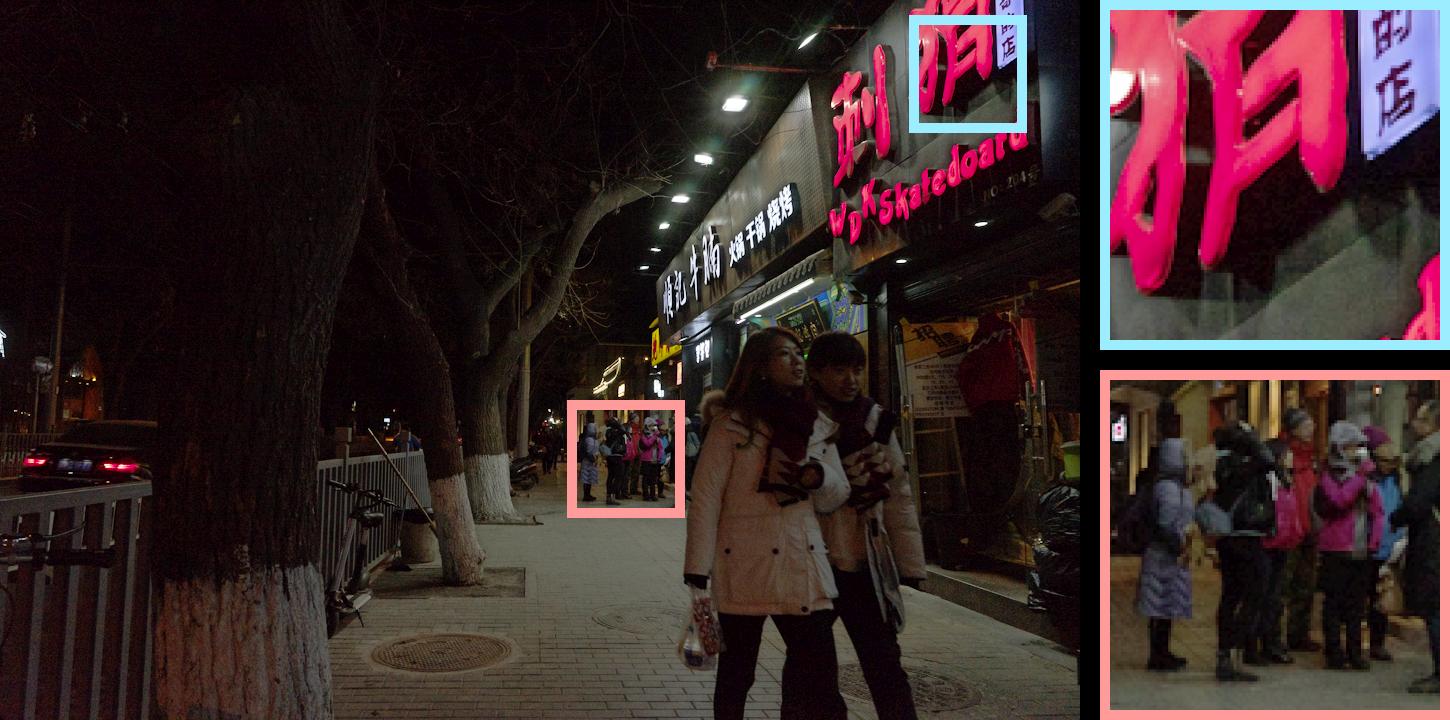}
        \caption{ZeroDCE$^{\bigtriangleup\bullet}$ \cite{guo2020zero}}
    \end{subfigure}%
    \newline
    \begin{subfigure}[b]{0.33\textwidth}
        \centering
        \includegraphics[width=.9\linewidth]{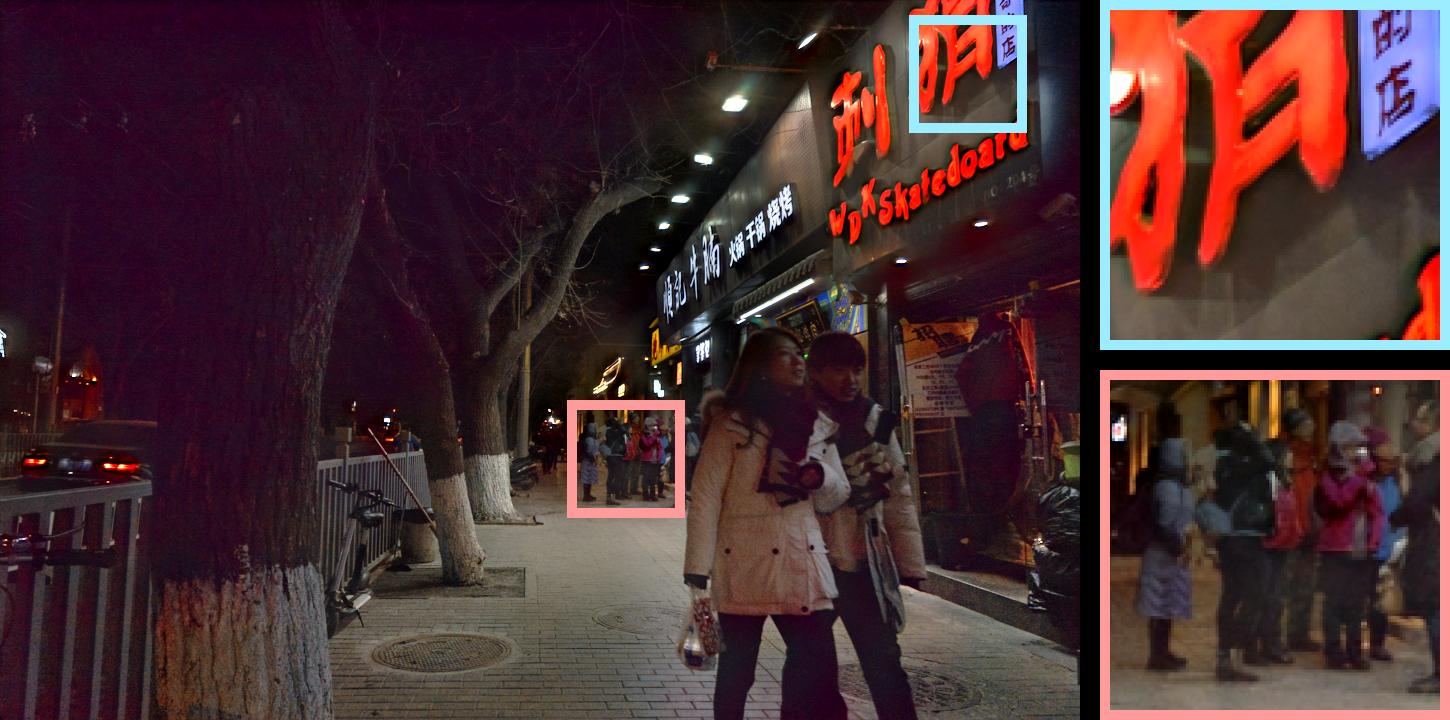}
        \caption{EnGAN$^\bigtriangleup$ \cite{jiang2021enlightengan}}
    \end{subfigure}%
    \hfill
    \begin{subfigure}[b]{0.33\textwidth}
        \centering
        \includegraphics[width=.9\linewidth]{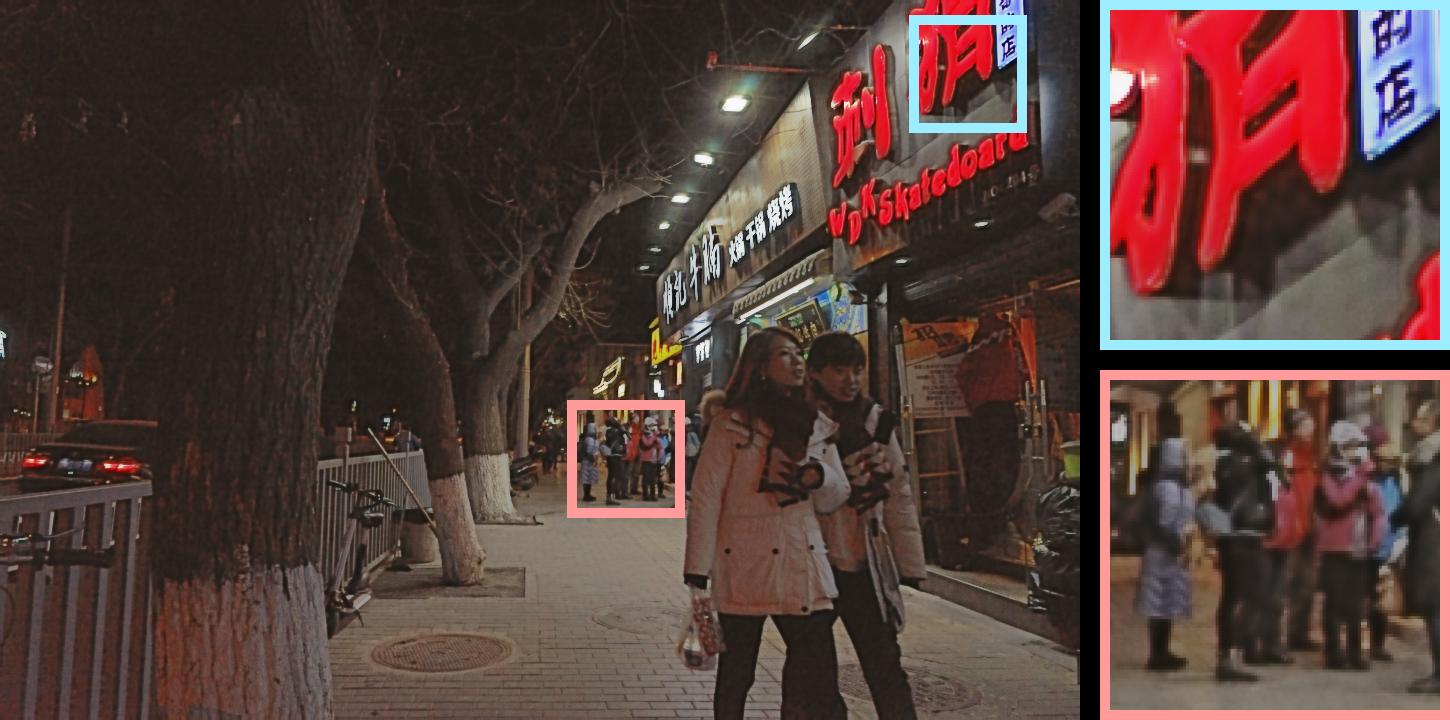}
        \caption{PairLIE$^\bigtriangleup$ \cite{fu2023learning}}
    \end{subfigure}%
    \hfill
    \begin{subfigure}[b]{0.33\textwidth}
        \centering
        \includegraphics[width=.9\linewidth]{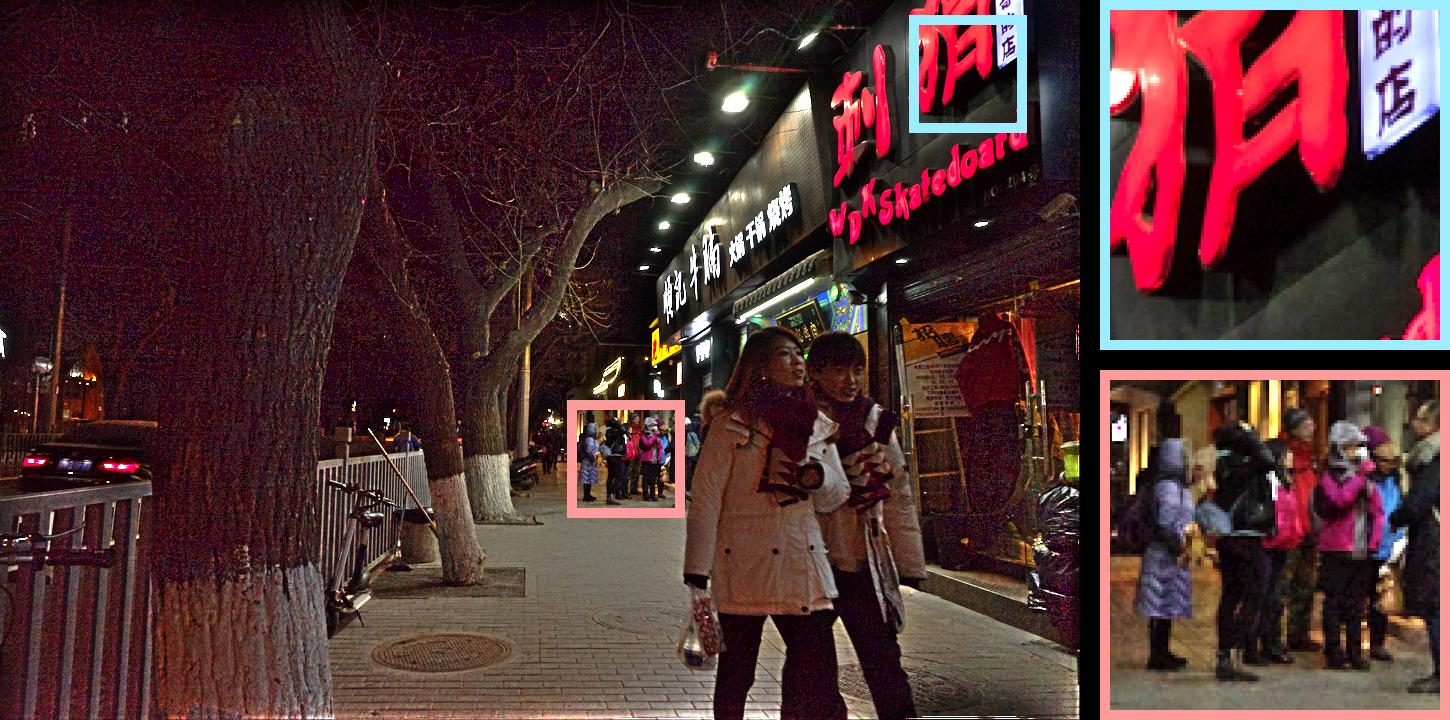}
        \caption{LIME$^\star$ \cite{guo2016lime}}
    \end{subfigure}%
    \newline
    \begin{subfigure}[b]{0.33\textwidth}
        \centering
        \includegraphics[width=.9\linewidth]{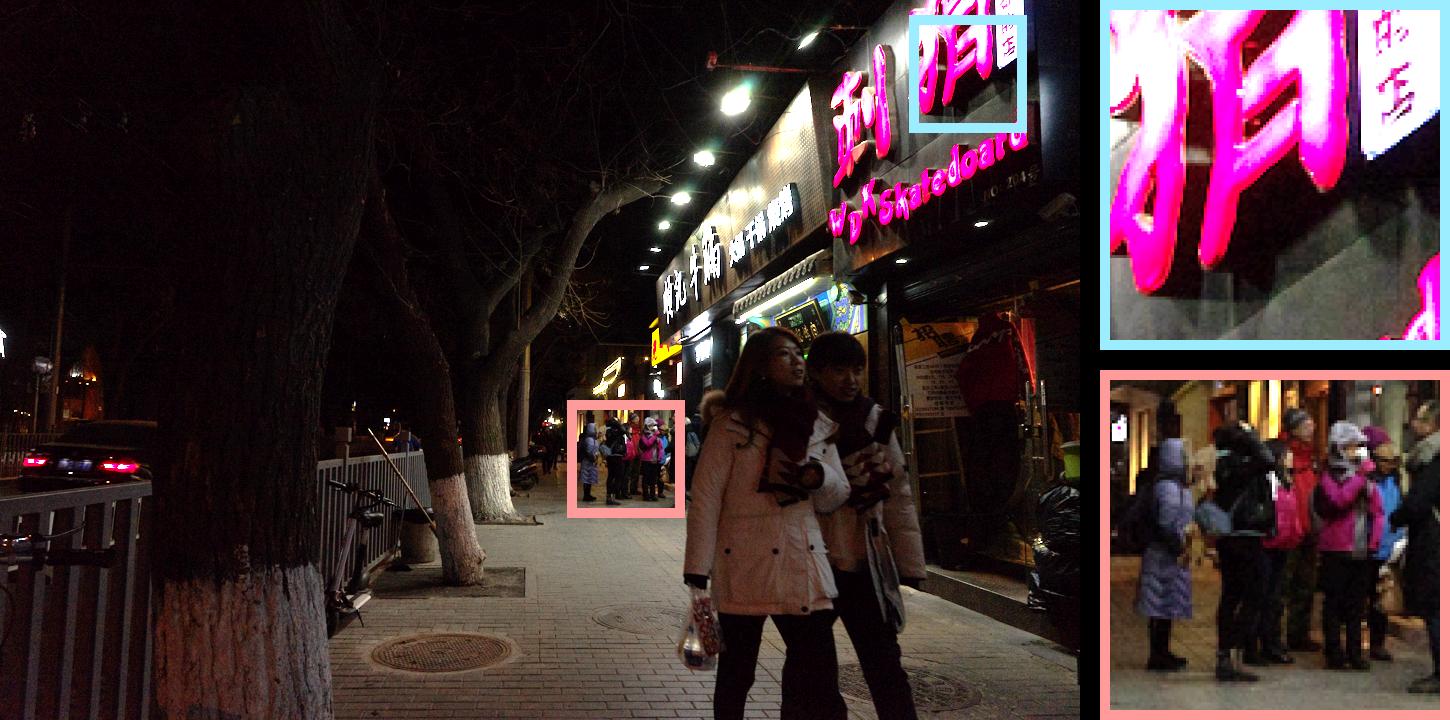}
        \caption{RUAS$^{\bigtriangleup\bullet}$ \cite{liu2021retinex}}
    \end{subfigure}%
    \hfill
    \begin{subfigure}[b]{0.33\textwidth}
        \centering
        \includegraphics[width=.9\linewidth]{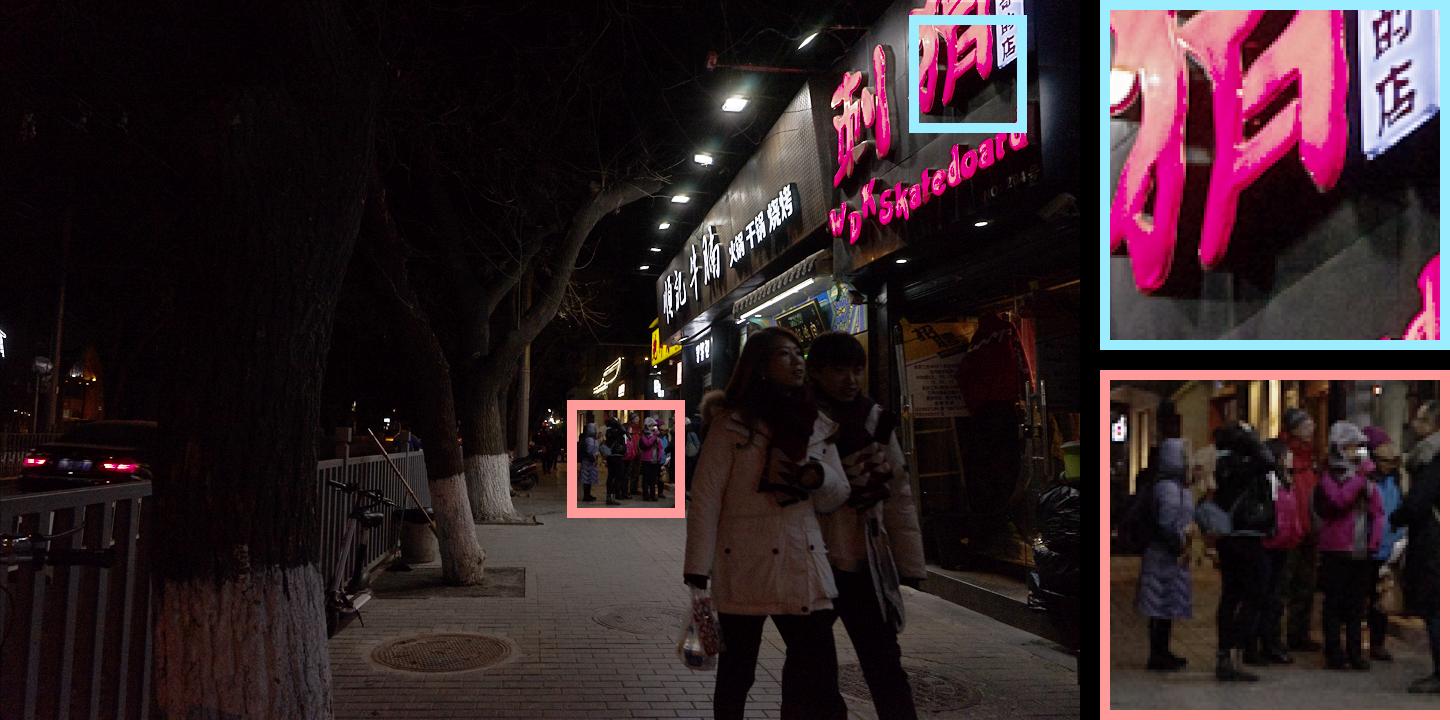}
        \caption{SCI$^{\bigtriangleup\bullet}$ \cite{ma2022toward}}
    \end{subfigure}%
    \hfill
    \begin{subfigure}[b]{0.33\textwidth}
        \centering
        \includegraphics[width=.9\linewidth]{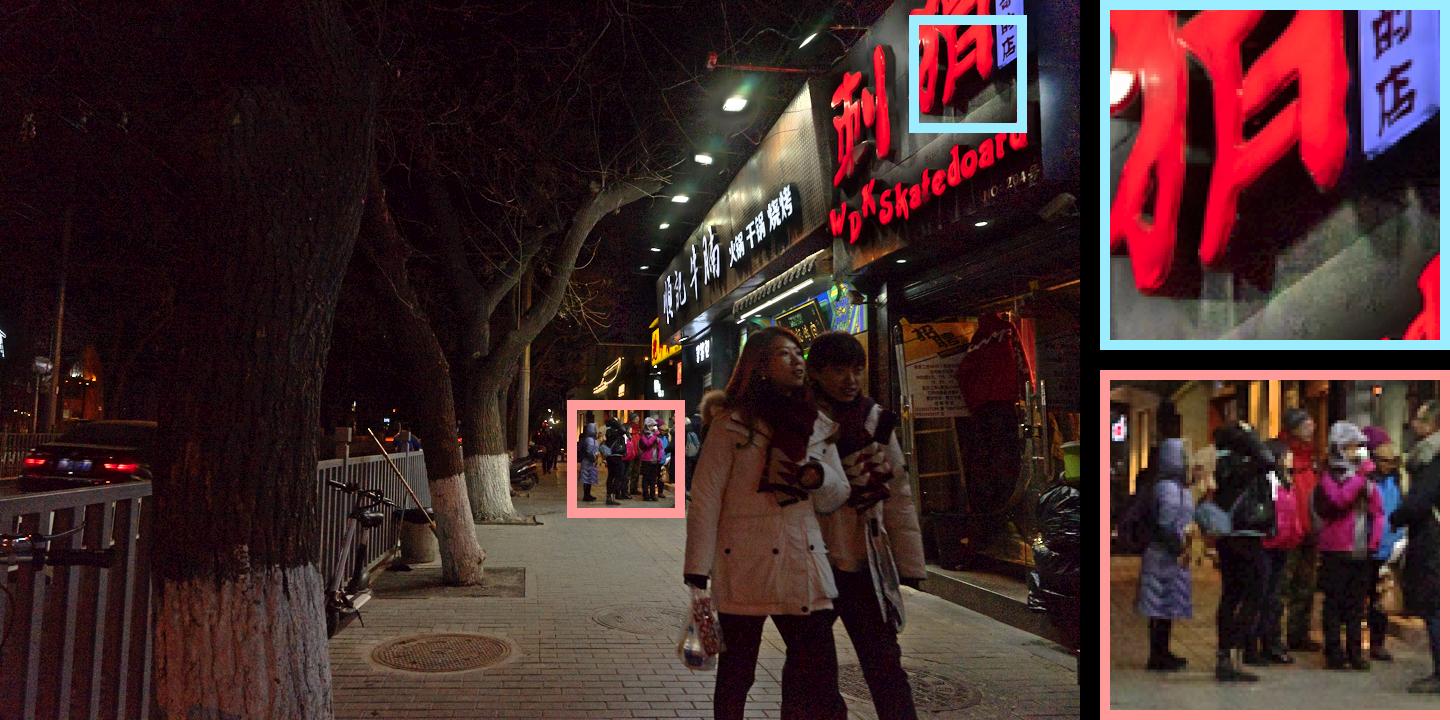}
        \caption{Ours$^\star$}
    \end{subfigure}%
    \caption{Visual comparison on a real-world night image from the DarkFace dataset. Labels $\dagger$, $\bigtriangleup$ and $\star$ denote supervised, unsupervised and zero-shot methods, respectively. Moreover, $\bullet$ indicates the method has been trained directly on images from the DarkFace dataset. Our method achieves the best perceptual quality with more stable improvement of under- and over-exposed regions without any color distortions.}\label{fig:darkface}
\end{figure}

Additionally, we provide a qualitative comparison using images from the MIT and DarkFace datasets in Figures \ref{fig:mit-qualitative} and \ref{fig:darkface}, respectively. Our proposed method exhibits superior performance by preserving the original colors of the image while efficiently enhancing the under-exposed regions. We also include quantitative no-reference evaluation of the enhancement measured by the entropy using the EMEE metric \cite{emee} in Table \ref{table:emee} given a random subset of 100 images from the DarkFace dataset.

\subsubsection{DarkFace Detection.} 
\begin{figure}[]
    \begin{subfigure}[b]{0.2\textwidth}
        \centering
        \includegraphics[width=0.8\linewidth]{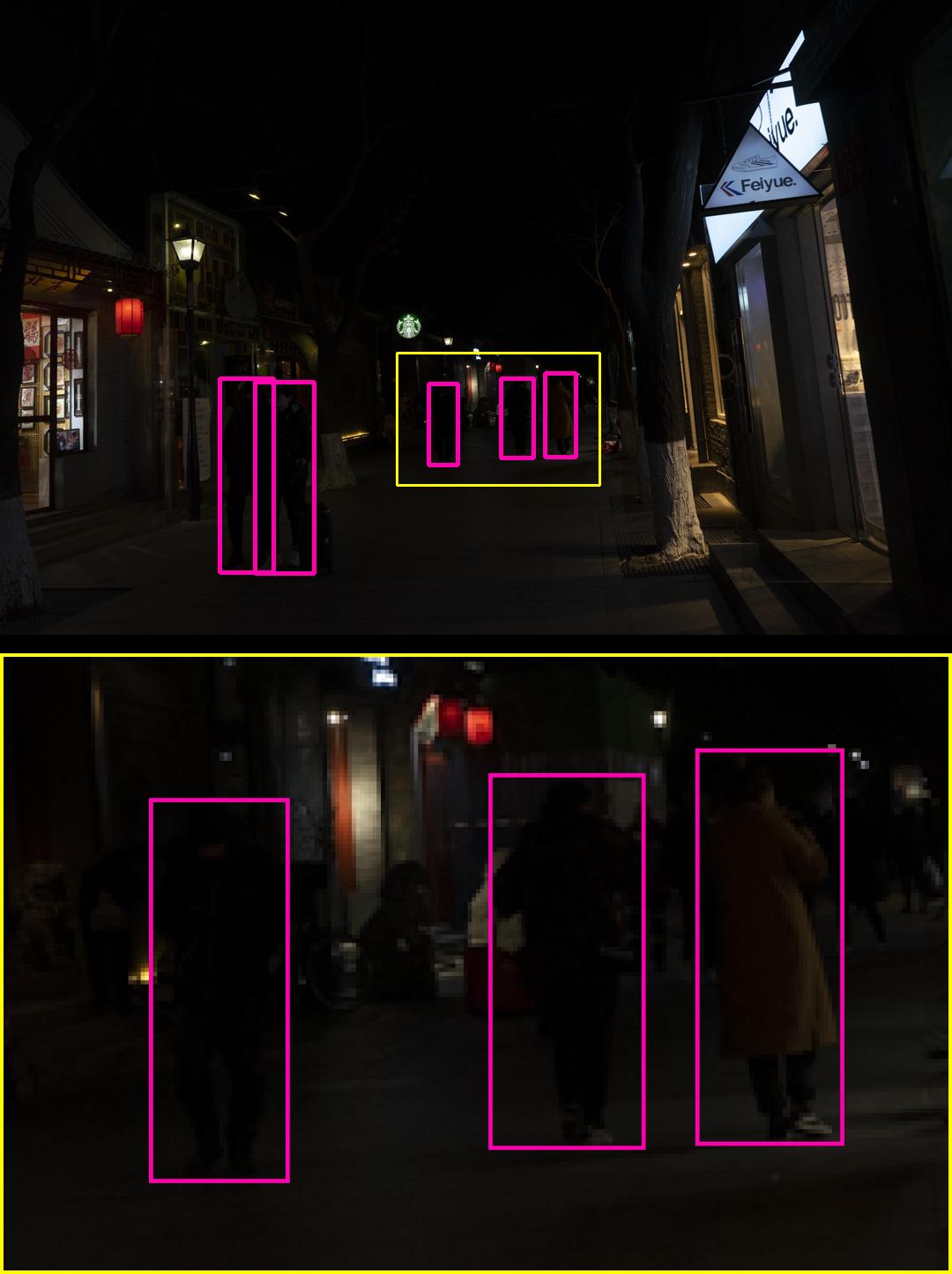}
        \caption{Input}
    \end{subfigure}%
    \hfill
    \begin{subfigure}[b]{0.2\textwidth}
        \centering
        \includegraphics[width=0.8\linewidth]{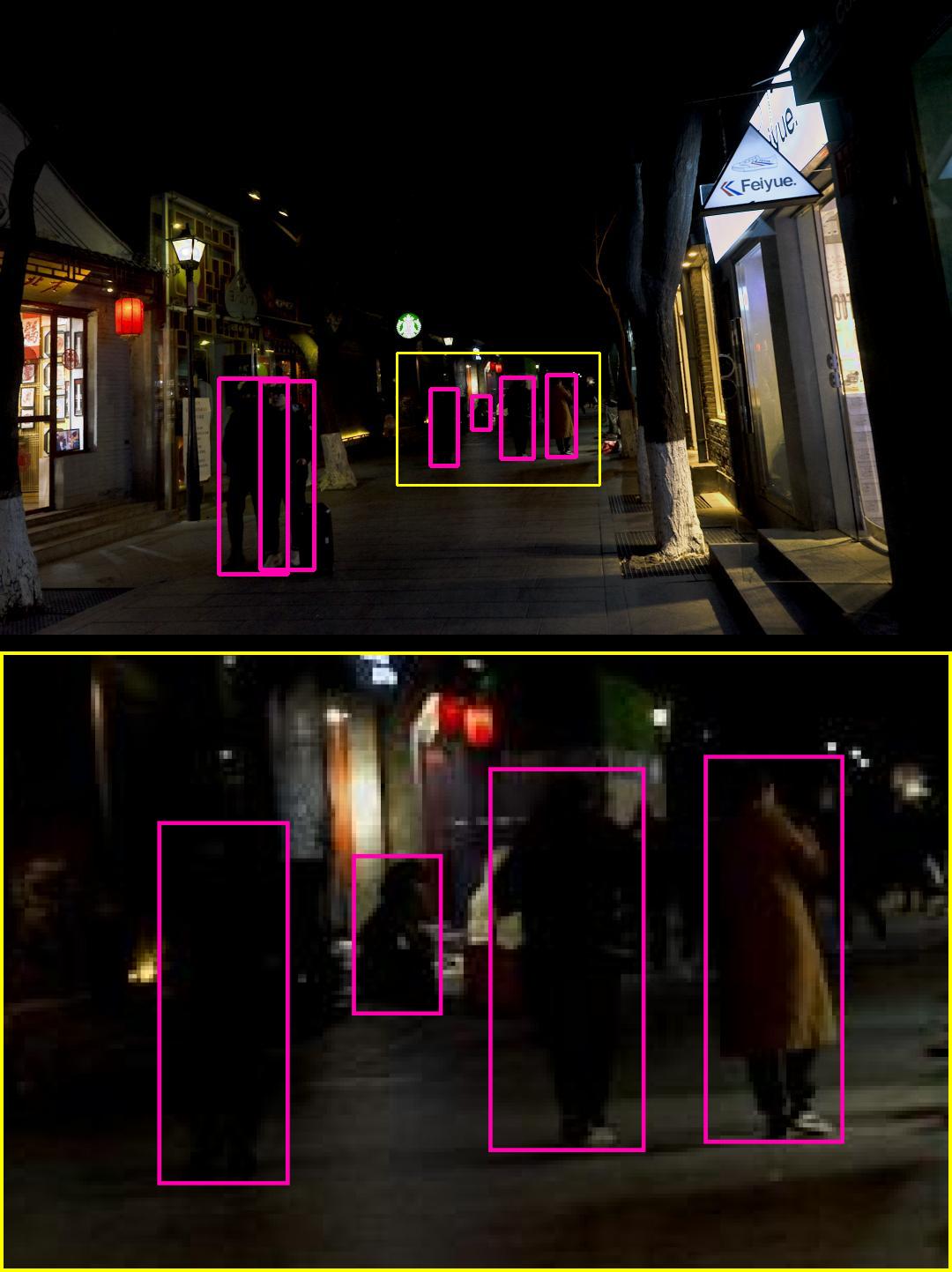}
        \caption{DeepLPF \cite{fan2022half}}
    \end{subfigure}%
    \hfill
    \begin{subfigure}[b]{0.2\textwidth}
        \centering
        \includegraphics[width=0.8\linewidth]{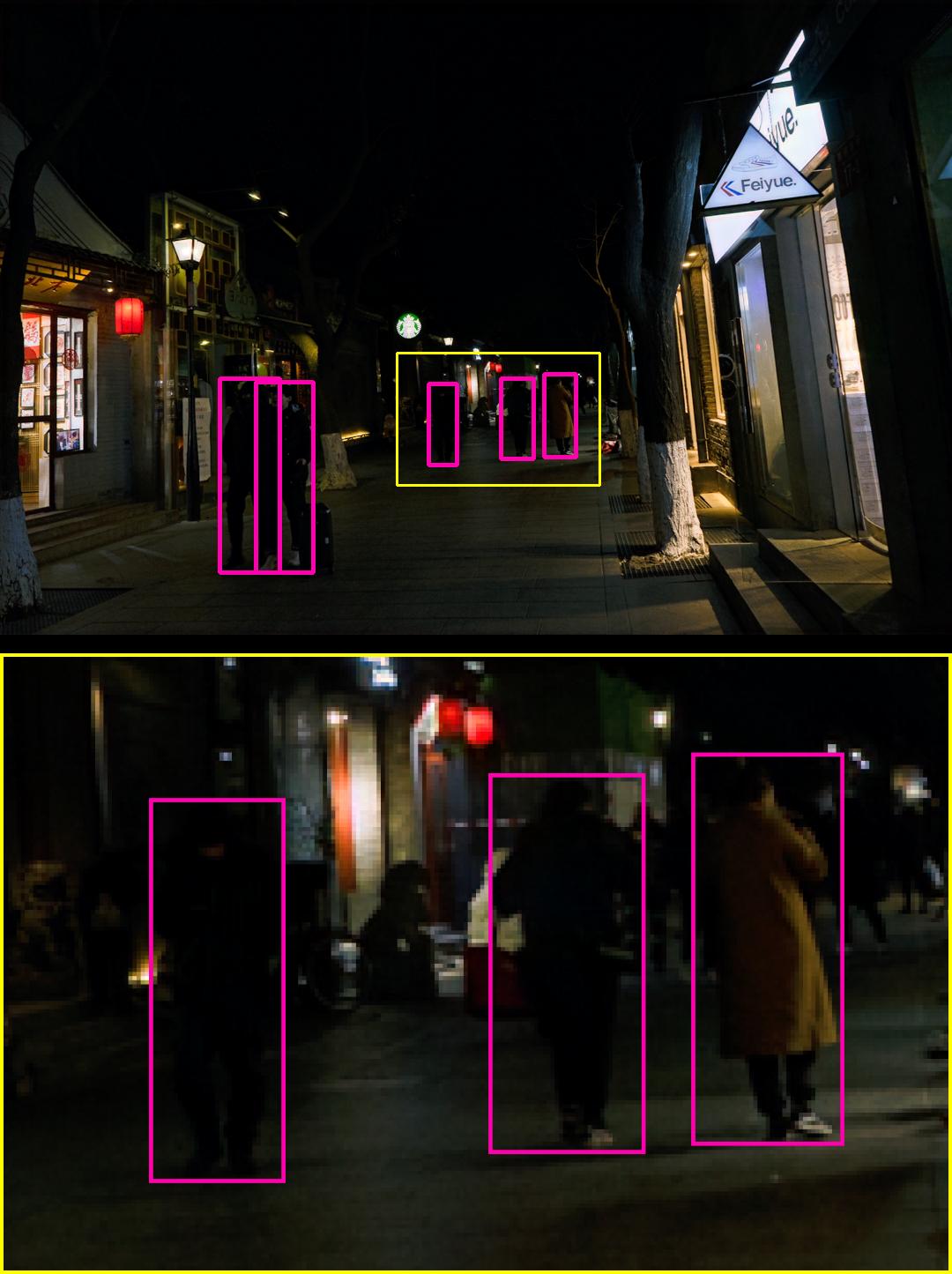}
        \caption{HWMNet \cite{fan2022half}}
    \end{subfigure}%
    \hfill
    \begin{subfigure}[b]{0.2\textwidth}
        \centering
        \includegraphics[width=0.8\linewidth]{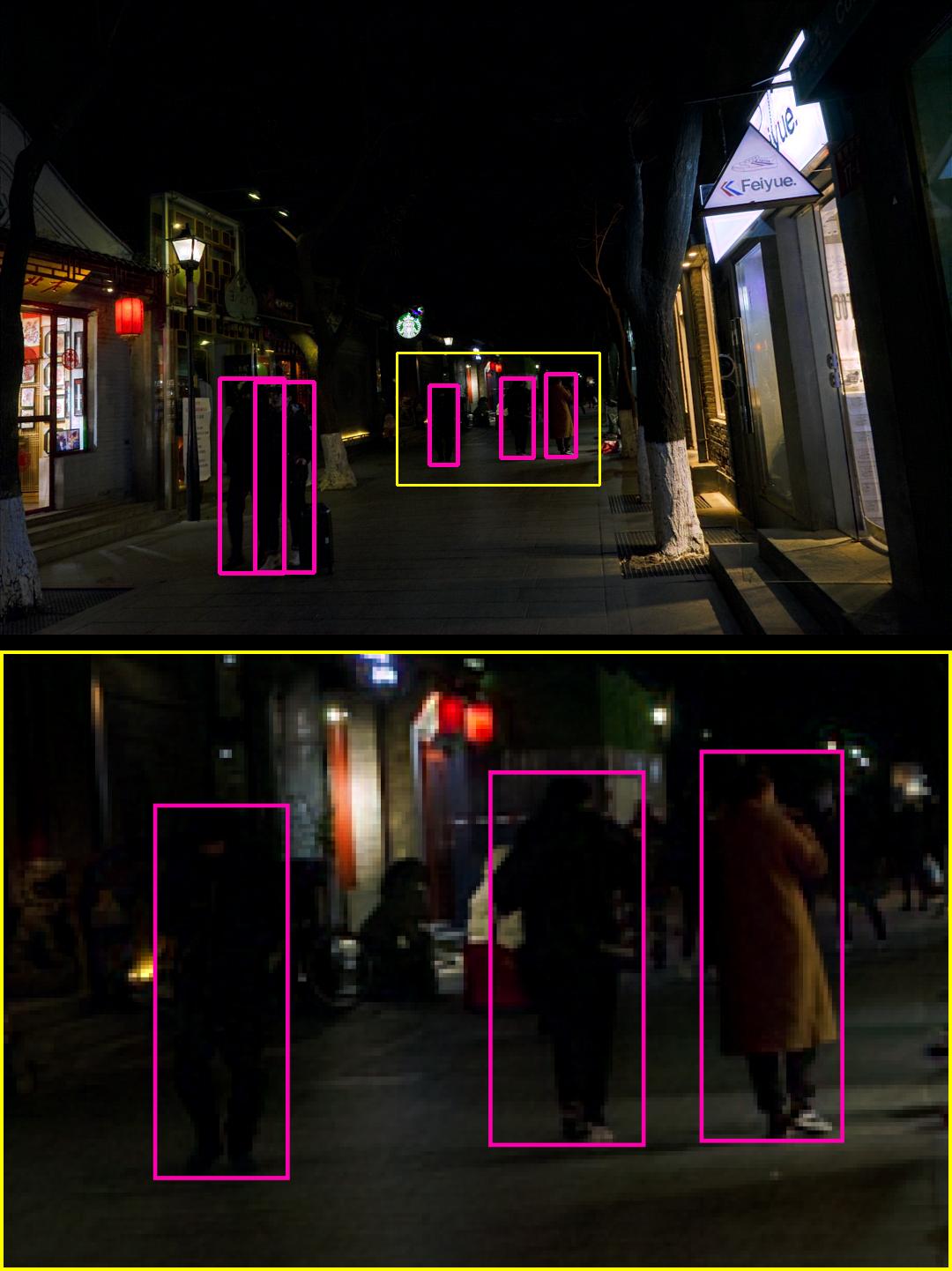}
        \caption{Zhao \textit{et al.} \cite{Zhao_2021_ICCV}}
    \end{subfigure}%
    \hfill
    \begin{subfigure}[b]{0.2\textwidth}
        \centering
        \includegraphics[width=0.8\linewidth]{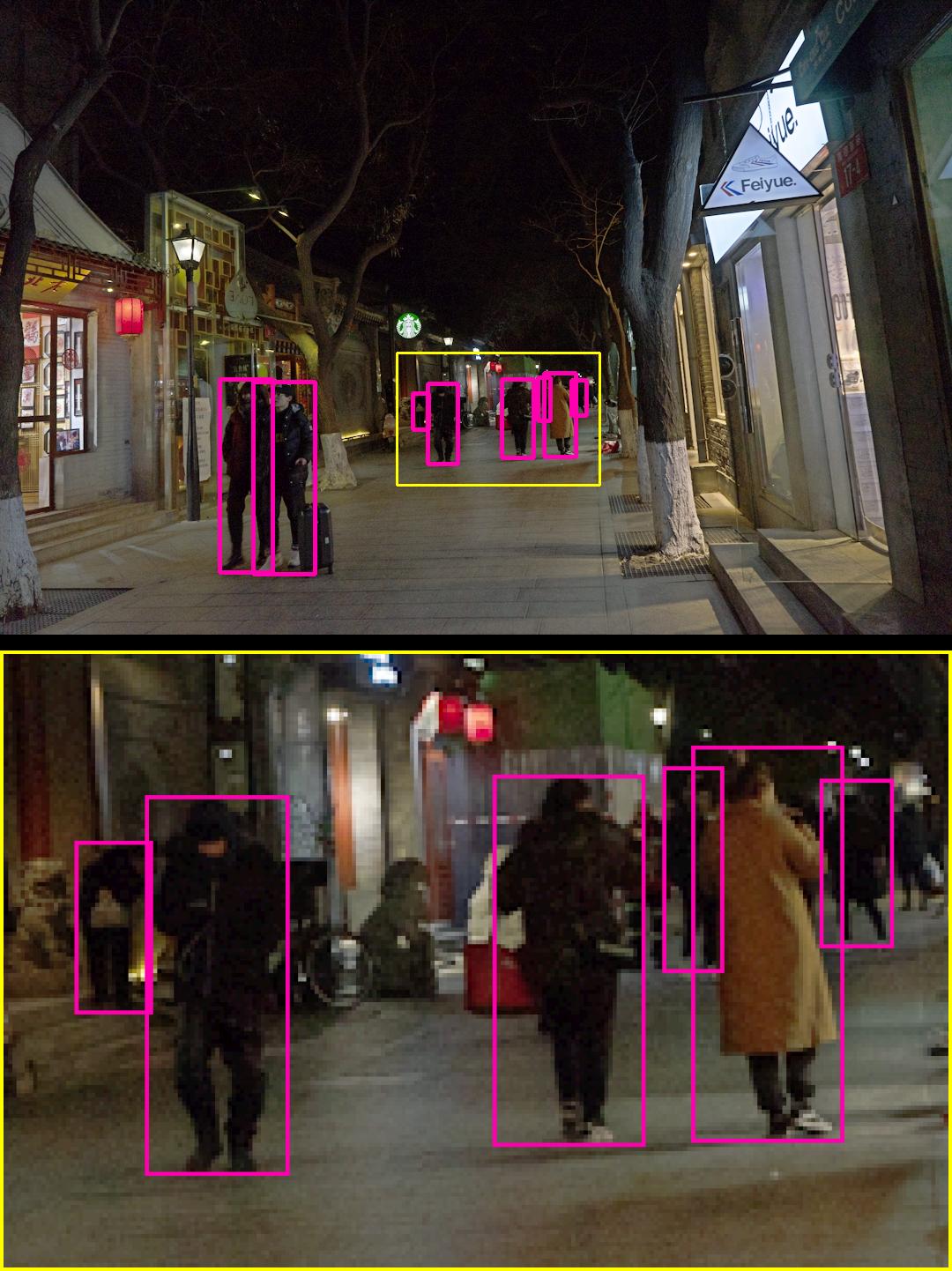}
        \caption{ZeroDCE \cite{guo2020zero}}
    \end{subfigure}%
    \newline
    \begin{subfigure}[b]{0.2\textwidth}
        \centering
        \includegraphics[width=0.8\linewidth]{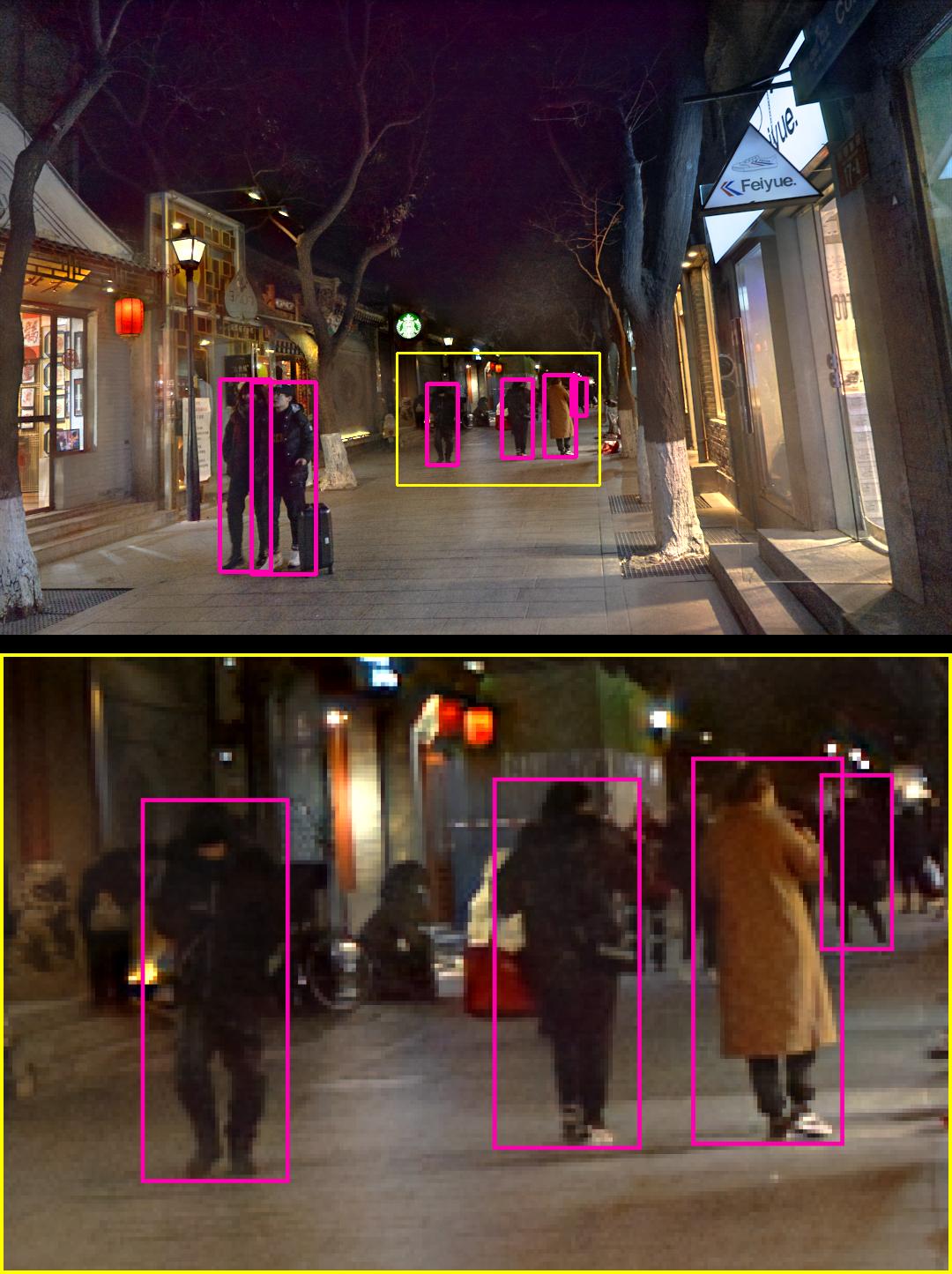}
        \caption{EnGAN \cite{jiang2021enlightengan}}
    \end{subfigure}%
    \hfill
    \begin{subfigure}[b]{0.2\textwidth}
        \centering
        \includegraphics[width=0.8\linewidth]{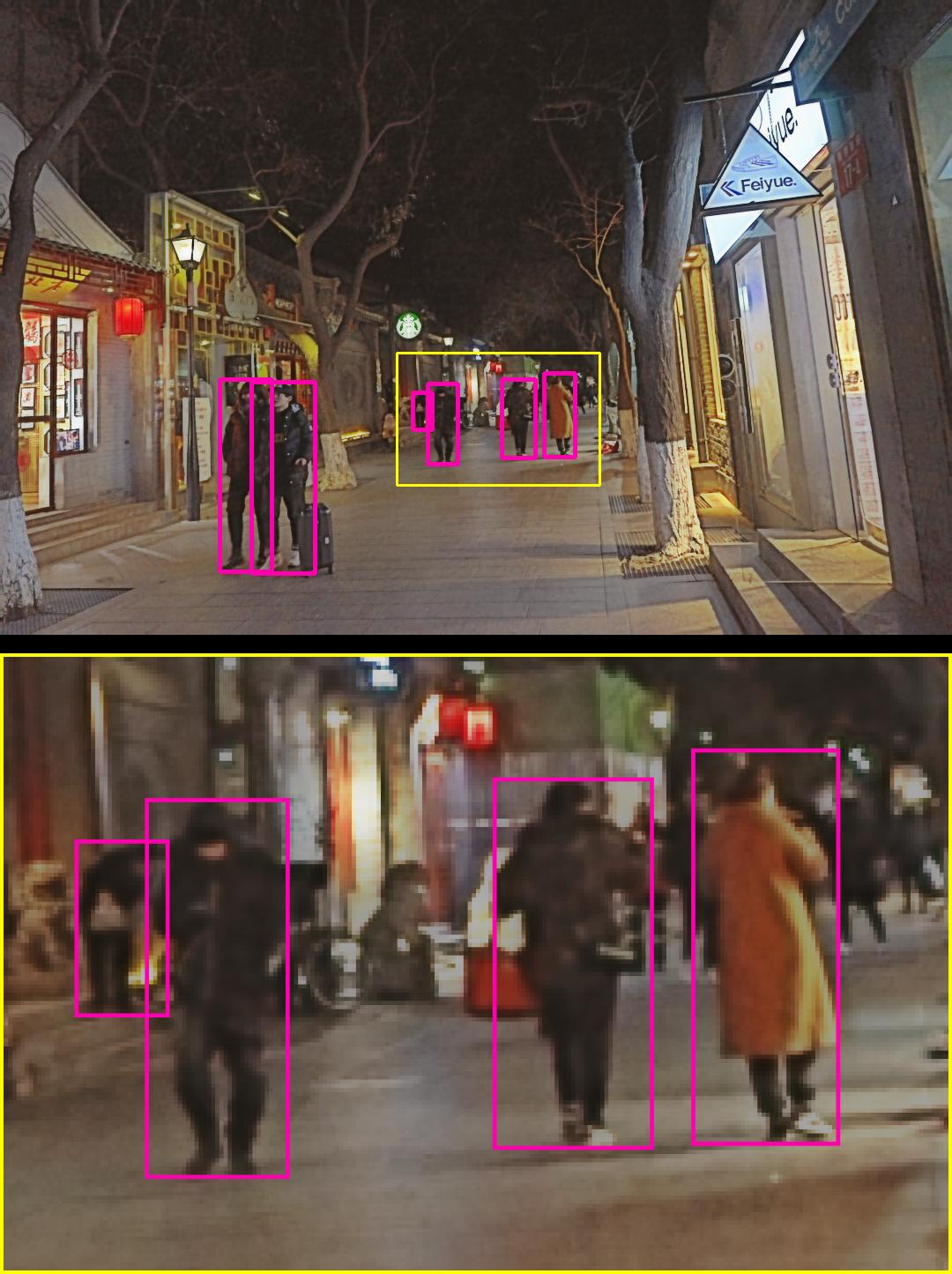}
        \caption{PairLIE \cite{fu2023learning}}
    \end{subfigure}%
    \hfill
    \begin{subfigure}[b]{0.2\textwidth}
        \centering
        \includegraphics[width=0.8\linewidth]{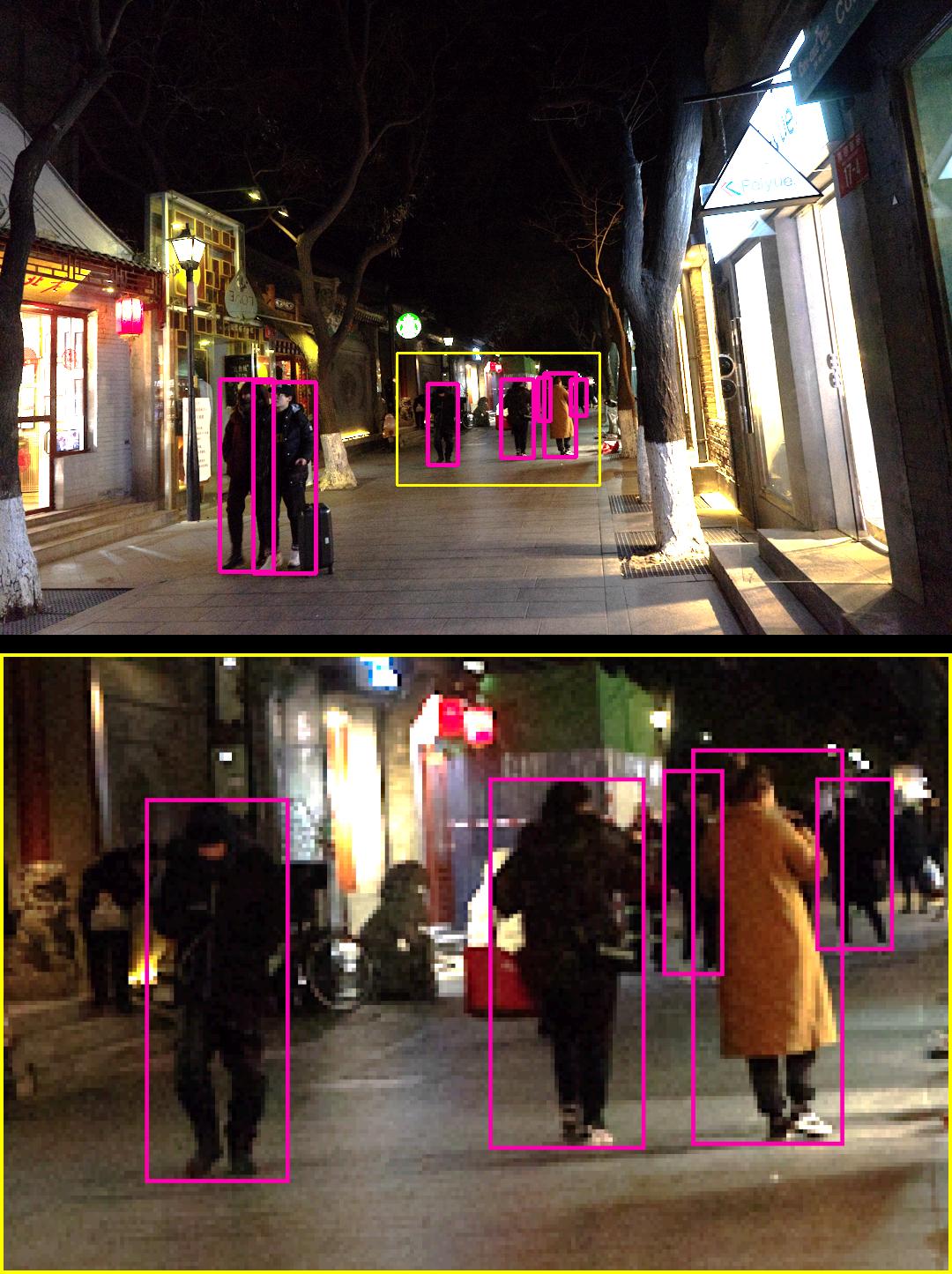}
        \caption{RUAS \cite{liu2021retinex}}
    \end{subfigure}%
    \hfill
    \begin{subfigure}[b]{0.2\textwidth}
        \centering
        \includegraphics[width=0.8\linewidth]{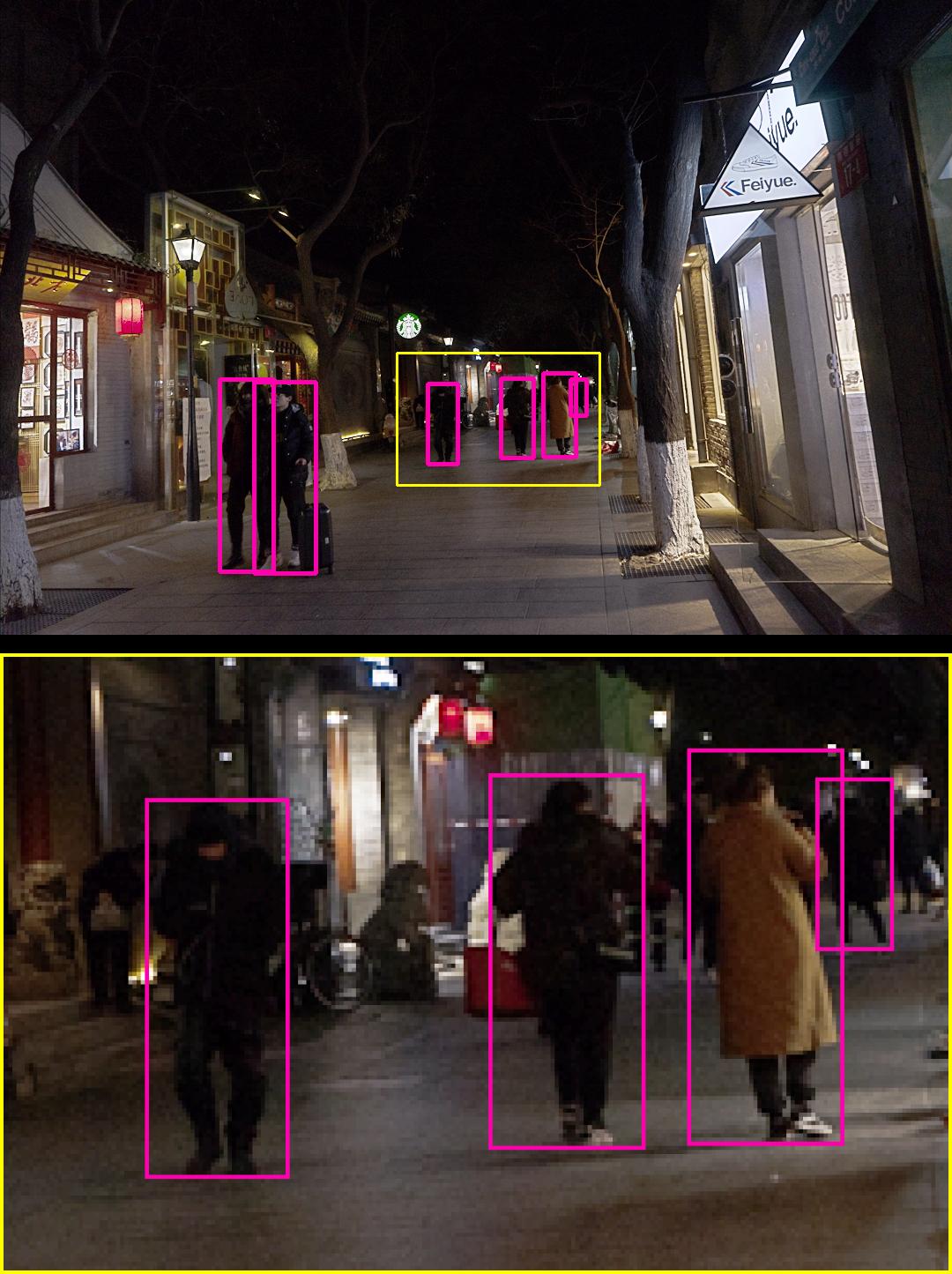}
        \caption{SCI \cite{ma2022toward}}
    \end{subfigure}%
    \hfill
    \begin{subfigure}[b]{0.2\textwidth}
        \centering
        \includegraphics[width=0.8\linewidth]{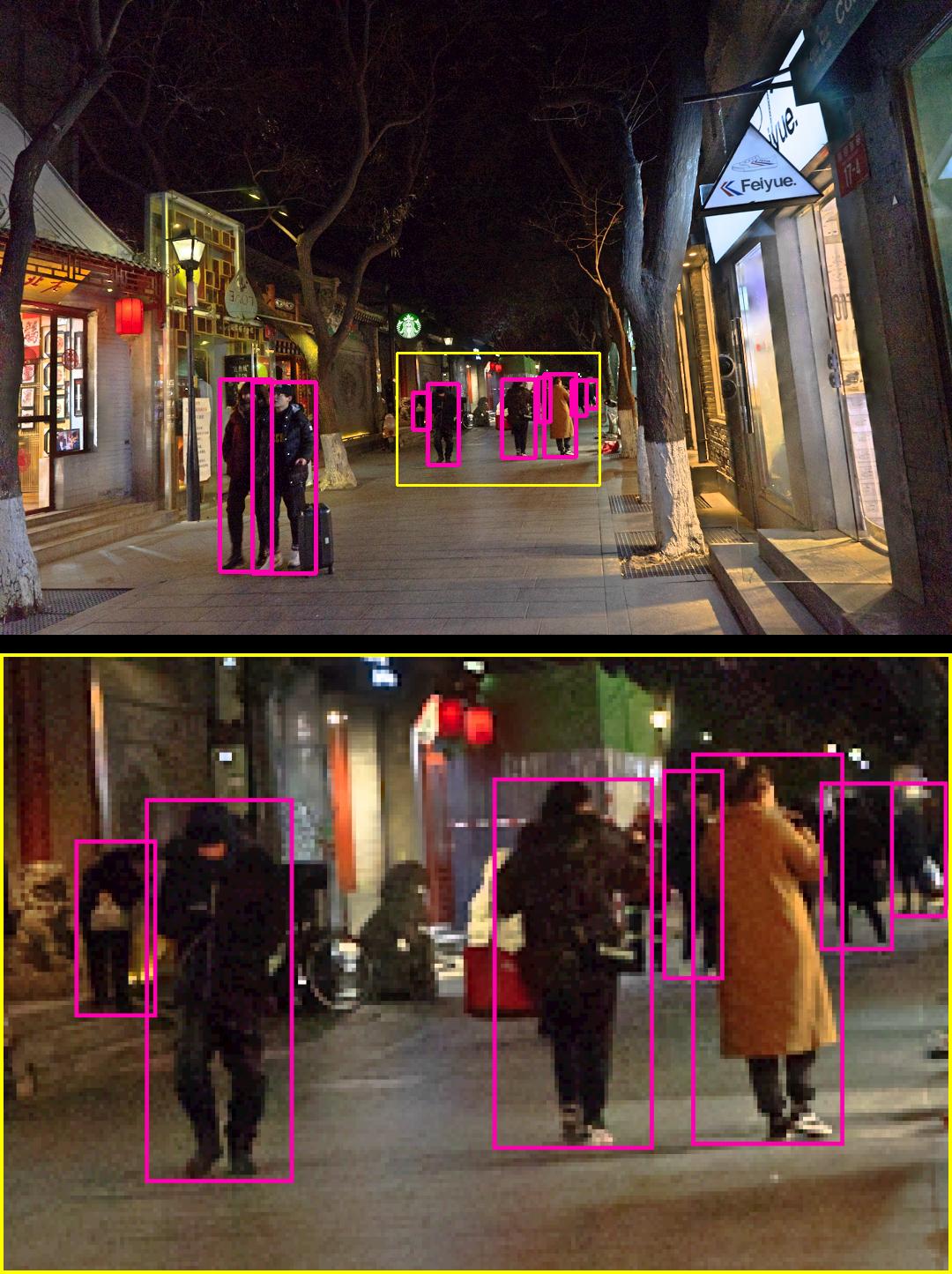}
        \caption{Ours}
    \end{subfigure}%
    \caption{An overview of the downstream detection task given an image from the DarkFace dataset. Our proposed CoLIE stands out as the only method capable of revealing all individuals present in the background, which are subsequently accurately detected.}\label{fig:detection}
\end{figure}

To assess the effectiveness of our method, we utilized Grounding DINO \cite{liu2023grounding} to detect individuals in images, utilizing the term "person" as the detection prompt. Our evaluation reveals that images enhanced by CoLIE exhibit heightened detection efficacy, resulting in an increased number of correctly detected objects as shown in Fig. \ref{fig:detection}. This indicates the enhanced performance of the detection model when integrating our method into image pre-processing pipelines.

\begin{table}[t]
\centering
\caption{Benchmarking study on the testing set of the UHD-LL dataset. We highlight the {\color{red}best} and the {\color{blue}second} and {\color{blue}third} best results.}
\label{table:uhd}
\scriptsize
\begin{tabular}{>{\centering\arraybackslash}p{3cm}||>{\centering\arraybackslash}p{2.5cm}||>{\centering\arraybackslash}p{1.5cm}|>{\centering\arraybackslash}p{1.5cm}||>{\centering\arraybackslash}p{2.5cm}}
Method & Training Setting & PSNR$\uparrow$ & SSIM$\uparrow$ & Training Set \\
\hline\hline
DeepLPF \cite{Moran_2020_CVPR}              & Supervised           & 16.3087       & 0.5322        & MIT-Adobe FiveK \\
\hline
HWMNet \cite{fan2022half}                   & Supervised           & 14.3063       & 0.5441        & MIT-Adobe FiveK \\
\hline
RetinexNet \cite{Chen2018Retinex}           & Supervised           & 16.0355       & 0.5796        & LOL \\
\hline
Zhao \textit{et al.} \cite{Zhao_2021_ICCV}  & Supervised           & 15.1646       & 0.5465        & MIT-Adobe FiveK \\
\hline\hline
ZeroDCE \cite{guo2020zero}                  & Unsupervised         & 12.6484       & 0.5961        & DarkFace \\
\hline
SCI-easy \cite{ma2022toward}                & Unsupervised         & 15.5433       & 0.6078        & MIT-Adobe FiveK \\
SCI-medium \cite{ma2022toward}              & Unsupervised         & 15.4666       & 0.6218        & LOL+LSRW \\
SCI-difficult \cite{ma2022toward}           & Unsupervised         & \noindent\color{blue}{17.8476}       & 0.5757        & DarkFace \\
\hline
RUAS-LOL \cite{liu2021retinex}              & Unsupervised         & 11.6932       & \noindent{\color{red}0.6970}        & LOL-v2 \\
RUAS-MIT5K \cite{liu2021retinex}            & Unsupervised         & 14.2146       & 0.5894        & MIT-Adobe FiveK \\
RUAS-DARK \cite{liu2021retinex}             & Unsupervised         & 11.2424       & 0.5737        & DarkFace \\
\hline
EnGAN \cite{jiang2021enlightengan}          & Unsupervised         & 13.1644       & 0.5964        & Assembled \\
\hline
PairLIE \cite{fu2023learning}               & Unsupervised         & \noindent{\color{blue}17.5349}       & \noindent{\color{blue}0.6847}        & LOL+SICE \\
\hline\hline
LIME \cite{guo2016lime}                     & Zero-shot            & 17.3496       & 0.5418        & -- \\
\hline
Ours                                        & Zero-shot            & \noindent{\color{red}17.8927}       & \noindent{\color{blue}0.6252}        & --                  
\end{tabular}
\end{table}

\begin{table}[t]
\centering
\caption{Benchmarking evaluation on a subset of the MIT dataset. We highlight the {\color{red}best} and the {\color{blue}second} and {\color{blue}third} best results.}
\label{table:mit}
\scriptsize
\begin{tabular}{>{\centering\arraybackslash}p{3cm}||>{\centering\arraybackslash}p{2.5cm}||>{\centering\arraybackslash}p{1.5cm}|>{\centering\arraybackslash}p{1.5cm}||>{\centering\arraybackslash}p{2.5cm}}
Method & Training Setting & PSNR$\uparrow$ & SSIM$\uparrow$ & Training Set \\
\hline\hline
HWMNet \cite{fan2022half}                   & Supervised           & \noindent\color{blue}{17.0284}       & \noindent\color{blue}{0.8190}        & LOL \\
\hline
RetinexNet \cite{Chen2018Retinex}           & Supervised           & 12.0806       & 0.6653        & LOL \\
\hline\hline
ZeroDCE \cite{guo2020zero}                  & Unsupervised         & \noindent\color{blue}{15.2448}       & 0.7509        & SICE \\
\hline
SCI-medium \cite{ma2022toward}              & Unsupervised         & 12.2299       & 0.7385        & LOL+LSRW \\
SCI-difficult \cite{ma2022toward}           & Unsupervised         & 15.0544       & \noindent\color{blue}{0.7941}        & DarkFace \\
\hline
RUAS-LOL \cite{liu2021retinex}              & Unsupervised         & 7.3771        & 0.4903        & LOL-v2 \\
RUAS-DARK \cite{liu2021retinex}             & Unsupervised         & 7.0460        & 0.4787        & DarkFace \\
\hline
EnGAN \cite{jiang2021enlightengan}          & Unsupervised         & 12.5576       & 0.7662        & Assembled \\
\hline
PairLIE \cite{fu2023learning}               & Unsupervised         & 12.8361       & 0.7677        & LOL+SICE \\
\hline\hline
LIME \cite{guo2016lime}                     & Zero-shot            & 14.9763       & 0.5418        & -- \\
\hline
Ours                                        & Zero-shot            & \noindent\color{red}{18.6703}       & \noindent\color{red}{0.8317}        & --                  
\end{tabular}
\end{table}

\begin{table}[t]
\centering
\caption{Benchmarking evaluation on a subset of images from the DarkFace dataset using a no-reference metric.}
\label{table:emee}
\scriptsize
\begin{tabular}{>{\centering\arraybackslash}p{3cm}||>{\centering\arraybackslash}p{2.5cm}||>{\centering\arraybackslash}p{1.5cm}}
Method & Training Setting & EMEE$\uparrow$ \\
\hline\hline
Input                                       & --                   & 2.6286 \\
\hline
DeepLPF \cite{Moran_2020_CVPR}              & Supervised           & 4.0142 \\
\hline
HWMNet \cite{fan2022half}                   & Supervised           & 4.3744 \\
\hline
ZeroDCE \cite{guo2020zero}                  & Unsupervised         & 5.5686 \\
\hline
SCI-difficult \cite{ma2022toward}           & Unsupervised         & 6.3856 \\
\hline
RUAS-DARK \cite{liu2021retinex}             & Unsupervised         & 6.2555 \\
\hline
EnGAN \cite{jiang2021enlightengan}          & Unsupervised         & 2.9794 \\
\hline
PairLIE \cite{fu2023learning}               & Unsupervised         & 1.7985 \\
\hline
Ours                                        & Zero-shot            & \noindent\color{red}{6.7473} \\         
\end{tabular}
\end{table}

\section{Ablation Study}
\label{sec:ablation}

\subsubsection{Context Window.} 
\begin{figure}[]
    \begin{subfigure}[b]{0.2\textwidth}
        \centering
        \includegraphics[width=0.9\linewidth]{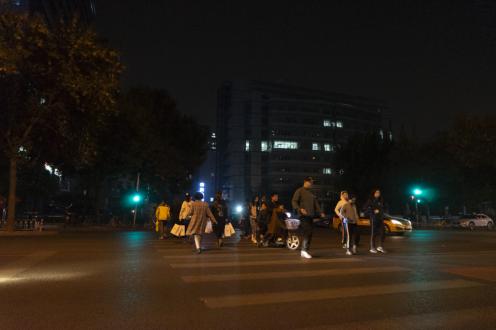}
        \caption{Input}
    \end{subfigure}%
    \begin{subfigure}[b]{0.2\textwidth}
        \centering
        \includegraphics[width=0.9\linewidth]{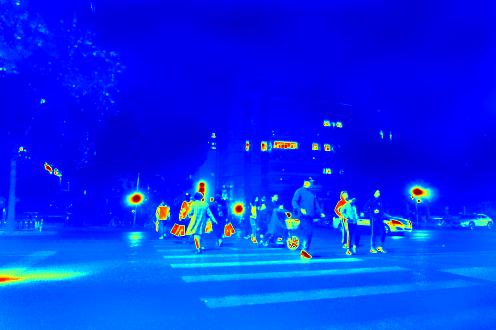}
        \caption{$1\times 1$}
    \end{subfigure}%
    \begin{subfigure}[b]{0.2\textwidth}
        \centering
        \includegraphics[width=0.9\linewidth]{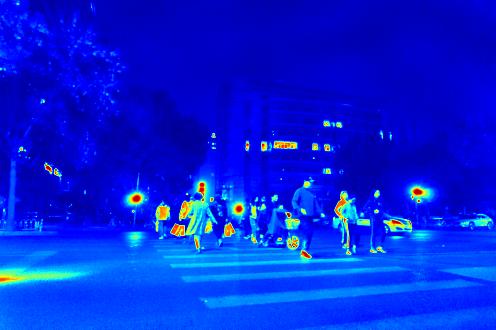}
        \caption{$3\times 3$}
    \end{subfigure}%
    \begin{subfigure}[b]{0.2\textwidth}
        \centering
        \includegraphics[width=0.9\linewidth]{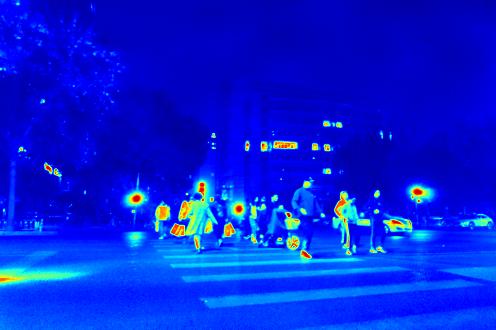}
        \caption{$5\times 5$}
    \end{subfigure}%
    \begin{subfigure}[b]{0.2\textwidth}
        \centering
        \includegraphics[width=0.9\linewidth]{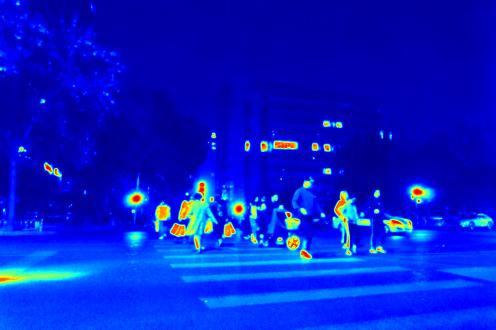}
        \caption{$7\times 7$}
    \end{subfigure}%
    \caption{Instances of predicted illumination components based on the size of the context window used for prediction (based on an image from the DarkFace dataset). Larger contextual windows output predictions of higher fidelity, particularly in areas with severe under-exposure.}\label{fig:ablation}
\end{figure}

We explore the impact of different context window sizes on the output of the model and show the results in Fig. \ref{fig:ablation}. It is evident that the model with a small window size of only one pixel encounters difficulties in accurately capturing image fidelity, especially in severely under-exposed regions. However, as the window size increases, the model captures more contextual information, resulting in finer, smoother, and more representative results. Moreover, the disparity in illumination strength between well-lit and dim regions becomes more pronounced with larger window sizes.

\subsubsection{Exposure Control.} The brightness level of the enhanced low-light image is primarily controlled by the parameter $L$ within the exposure term $\mathcal{L}_\textit{exp}$ of the loss function. Lower values of this parameter force the model to refine the illumination field more intensely, resulting in increased brightness and enhanced visibility of objects within the image. However, if the parameter is set too low, it may lead to over-exposure of the image, revealing noisy structures or introducing unnatural visual characteristics. Conversely, maintaining the parameter at high values can diminish the enhancement effect, retaining illumination levels similar to those of the input image. We present the results corresponding to various values of the parameter $L$ in Fig. \ref{fig:ablation-loss}.

\subsubsection{Contribution of Each Loss.} 
\begin{figure}[]
    \begin{subfigure}[b]{0.2\textwidth}
        \centering
        \includegraphics[width=0.9\linewidth]{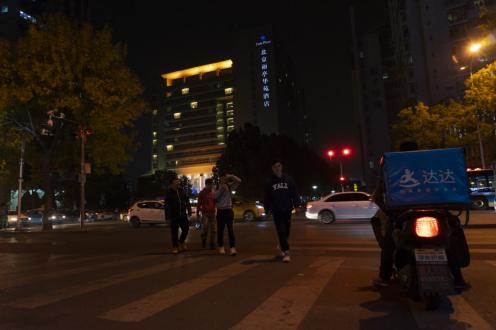}
        \caption{Input}
    \end{subfigure}%
    \begin{subfigure}[b]{0.2\textwidth}
        \centering
        \includegraphics[width=0.9\linewidth]{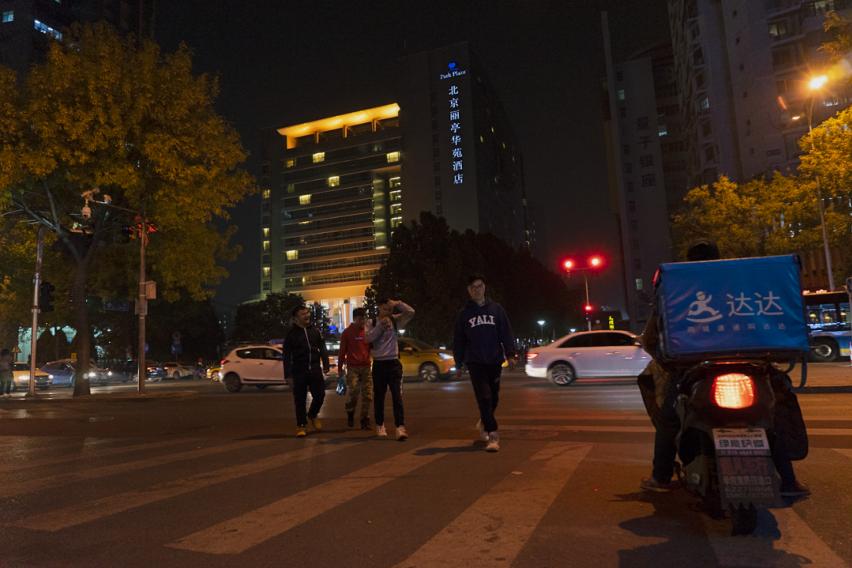}
        \caption{$L=0.9$}
    \end{subfigure}%
    \begin{subfigure}[b]{0.2\textwidth}
        \centering
        \includegraphics[width=0.9\linewidth]{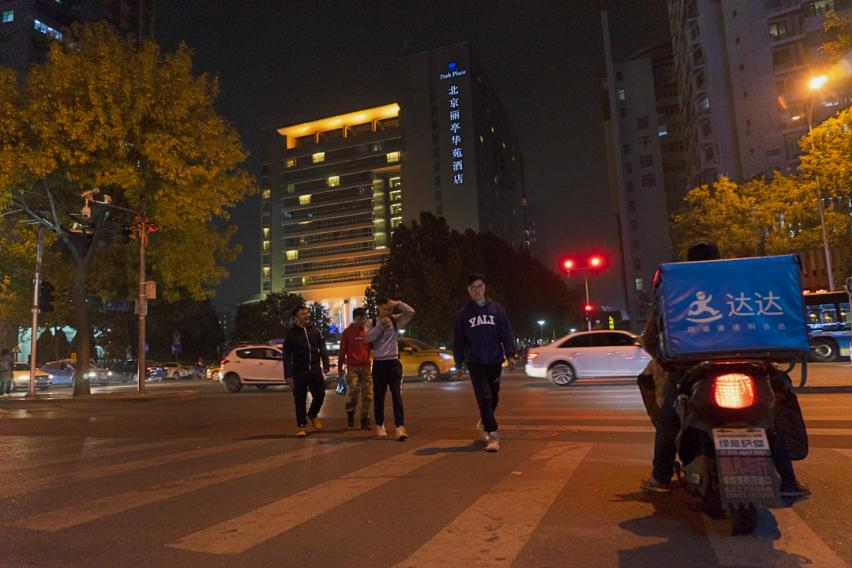}
        \caption{$L=0.7$}
    \end{subfigure}%
    \begin{subfigure}[b]{0.2\textwidth}
        \centering
        \includegraphics[width=0.9\linewidth]{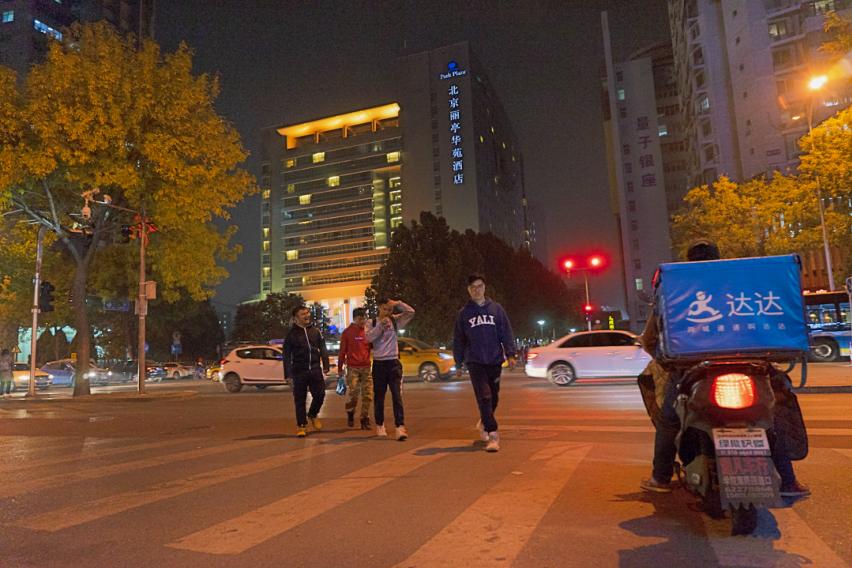}
        \caption{$L=0.5$}
    \end{subfigure}%
    \begin{subfigure}[b]{0.2\textwidth}
        \centering
        \includegraphics[width=0.9\linewidth]{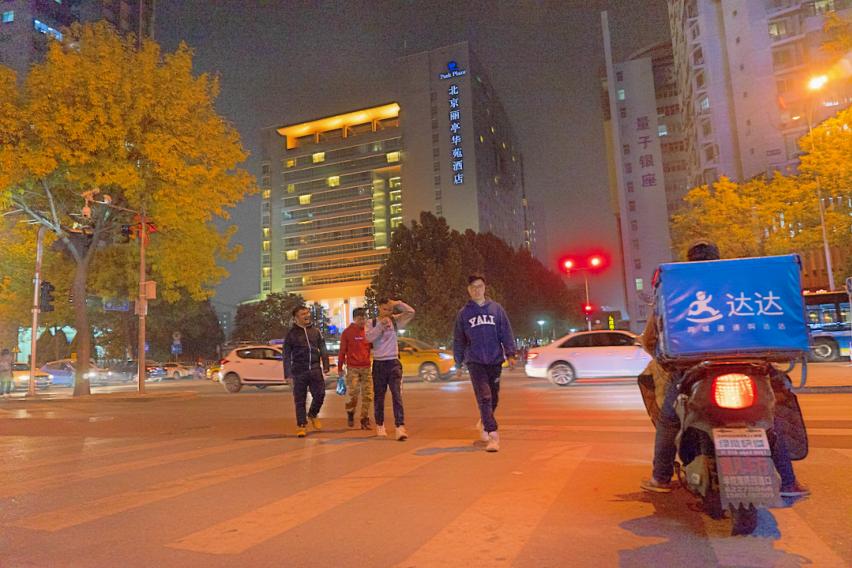}
        \caption{$L=0.3$}
    \end{subfigure}%
    \caption{Overview of the effect of the parameter $L$ in the exposure loss term on the brightness of the output image (based on an image from the DarkFace dataset).}\label{fig:ablation-loss}
\end{figure}

We present the quantitative results of our proposed approach applied to the MIT dataset based on various combinations of losses in Table \ref{table:loss}. Notably, our results demonstrate a substantial impact when the term enforcing sparsity, denoted as $\mathcal{L}_\textit{spa}$, is excluded. This particular term plays a critical role in ensuring that the enhanced images avoid over-exposure, thereby preserving high perceptual quality. On the contrary, the influence of the term $\mathcal{L}_\textit{exp}$, which regulates the level of brightness enhancement in the image, is less pronounced in quantitative evaluations. However, its inclusion remains important for maintaining image quality, particularly in low-light scenarios, as depicted in Fig. \ref{fig:ablation-loss}. Lastly, the omission of the illumination smoothness loss $\mathcal{L}_\textit{s}$ disrupts the correlations among neighboring regions, thereby violating the assumptions of the Retinex theory and consequently leading to a decline in performance.

\begin{table}[t]
\centering
\caption{Ablation study of loss variations based on the MIT dataset.}
\label{table:loss}
\begin{tabular}{>{\centering\arraybackslash}p{1.5cm}|>{\centering\arraybackslash}p{1.5cm}||>{\centering\arraybackslash}p{1.5cm}|>{\centering\arraybackslash}p{1.5cm}|>{\centering\arraybackslash}p{1.5cm}|>{\centering\arraybackslash}p{1.5cm}}
PSNR$\uparrow$ & SSIM$\uparrow$ & $\mathcal{L}_\textit{f}$ & $\mathcal{L}_\textit{s}$ & $\mathcal{L}_\textit{exp}$ & $\mathcal{L}_\textit{spa}$ \\
\hline\hline
17.3028 & 0.7891 & \checkmark & & \checkmark & \checkmark \\
15.5348 & 0.8132 & \checkmark & \checkmark & & \checkmark \\
 7.2470 & 0.4915 & \checkmark & \checkmark & \checkmark & \\
18.6703 & 0.8317 & \checkmark & \checkmark & \checkmark & \checkmark \\
\end{tabular}
\end{table}

\subsubsection{Microscopy Image Application.} 
\begin{figure}[ht]
    \centering
    \includegraphics[width=.7\linewidth]{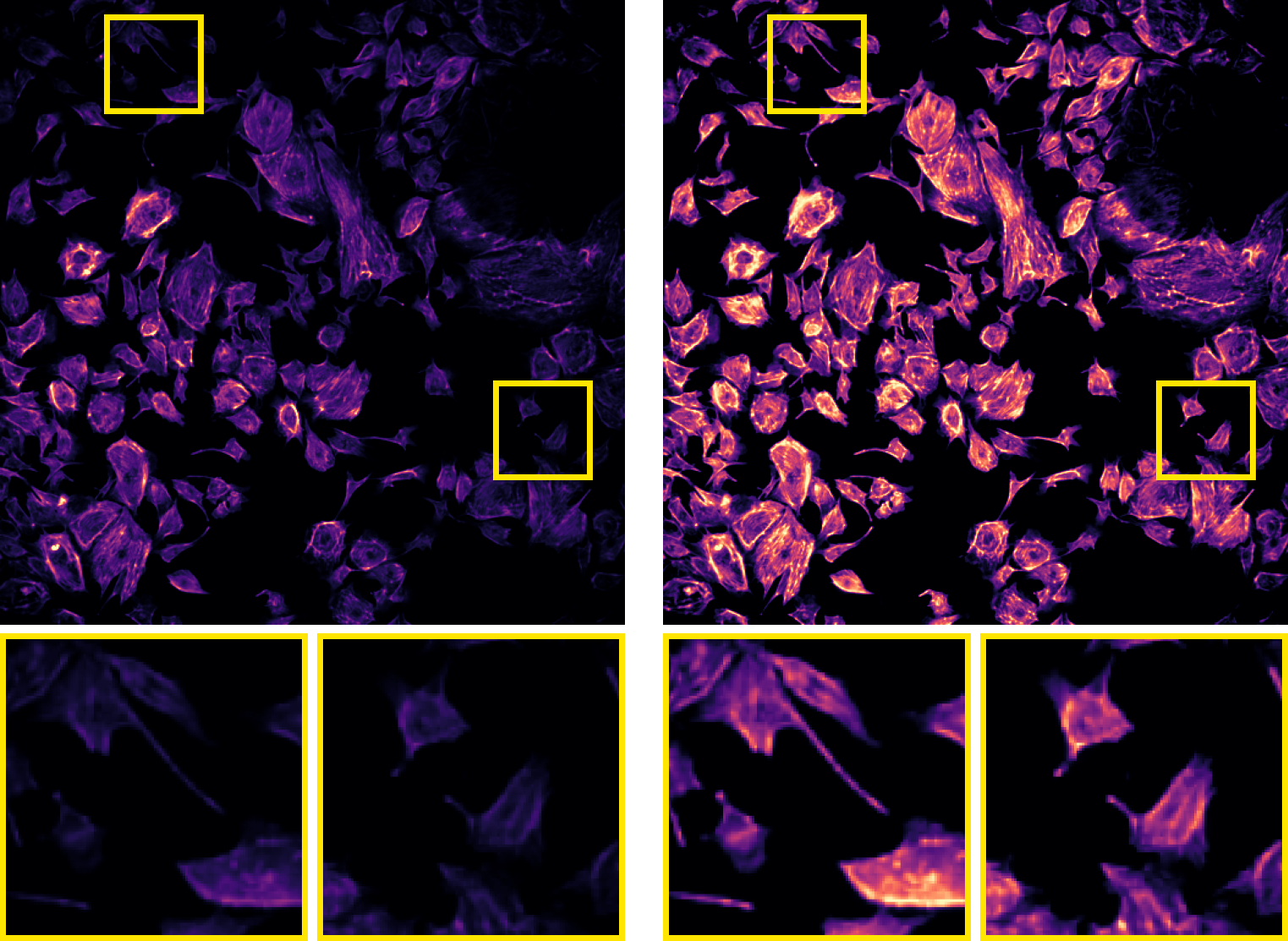}
    \caption{Visual example of a fluorescence microscopy image (left) and the enhancement results with CoLIE (right). The restoration process rectifies illumination irregularities and effectively eliminates vignetting, which are typical characteristics of fluorescence microscopy imaging.}\label{fig:microscopy}
\end{figure}

Optical microscopy data, especially fluorescence imaging, is often severely affected by shading or vignetting, as detailed by Goldman \cite{goldman2010vignette}. This issue is typically manifested as an attenuation of brightness intensity from the center of the optical axis to the edges. This not only degrades the visual quality of the image but, more critically, distorts the quantification of fluorescence intensities that are indicative of biological or molecular properties. While the existing microscopy intensity correction methodologies \cite{Smith2015, Peng2017} predominantly hinges on utilizing a series of images with a consistent illumination profile, their applicability to individual images remains limited. In this context, we apply our proposed CoLIE to a singular fluorescence image. Our findings underscore a significant enhancement in image intensity, with a pronounced improvement observable at the peripheries where shading effects are most acute (see Fig. \ref{fig:microscopy}).

\subsubsection{Limitations \& Discussion.} While our proposed method exhibits superior performance compared to previous approaches, it is important to note that there remain some open challenges it does not entirely address. Our method primarily targets the enhancement of the Value component within the HSV image representation. However, this focus introduces a vulnerability to noise, as each component of the HSV representation can be affected differently (see Fig. \ref{fig:hsv}). While our approach effectively improves the overall quality of images by manipulating the Value component, it does not fully mitigate noise interference originating from variations in the Hue and Saturation components. 

\section{Conclusion}

In this paper, we introduce CoLIE, a novel zero-shot low-light image enhancement model utilizing a neural implicit representation function (NIR). Our approach focuses on restoring the illumination component in accordance with the Retinex theory, aiming to enhance image quality and perceptual fidelity. Particularly, we integrate information from both image coordinates and input pixel values, beyond conventional NIR methods. Experimental results demonstrate superior performance on real-world night and under-exposed images, supported by both quantitative metrics and qualitative evaluations.

\section*{Acknowledgements}
Tomáš Chobola is supported by the Helmholtz Association under the joint research school "Munich School for Data Science - MUDS".

\clearpage  

%
%
\bibliographystyle{splncs04}
\bibliography{main}
\end{document}